# SATzilla: Portfolio-based Algorithm Selection for SAT

**Lin Xu**                                                    XULIN730@CS.UBC.CA
**Frank Hutter**                                              HUTTER@CS.UBC.CA
**Holger H. Hoos**                                            HOOS@CS.UBC.CA
**Kevin Leyton-Brown**                                        KEVINLB@CS.UBC.CA
*Department of Computer Science*
*University of British Columbia*
*201-2366 Main Mall, BC V6T 1Z4, CANADA*

## Abstract

It has been widely observed that there is no single "dominant" SAT solver; instead, different solvers perform best on different instances. Rather than following the traditional approach of choosing the best solver for a given class of instances, we advocate making this decision online on a per-instance basis. Building on previous work, we describe SATzilla, an automated approach for constructing per-instance algorithm portfolios for SAT that use so-called empirical hardness models to choose among their constituent solvers. This approach takes as input a distribution of problem instances and a set of component solvers, and constructs a portfolio optimizing a given objective function (such as mean runtime, percent of instances solved, or score in a competition). The excellent performance of SATzilla was independently verified in the 2007 SAT Competition, where our SATzilla07 solvers won three gold, one silver and one bronze medal. In this article, we go well beyond SATzilla07 by making the portfolio construction scalable and completely automated, and improving it by integrating local search solvers as candidate solvers, by predicting performance score instead of runtime, and by using hierarchical hardness models that take into account different types of SAT instances. We demonstrate the effectiveness of these new techniques in extensive experimental results on data sets including instances from the most recent SAT competition.

## 1. Introduction

The propositional satisfiability problem (SAT) is one of the most fundamental problems in computer science. Indeed, entire conferences and journals are devoted to the study of this problem, and it has a long history in AI. SAT is interesting both for its own sake and because instances of other problems in $\mathcal{NP}$ can be encoded into SAT and solved by SAT solvers. This approach has proven effective for tackling many real-world applications, including planning, scheduling, graph colouring, bounded model checking, and formal verification (examples have been described by Kautz & Selman, 1996, 1999; Crawford & Baker, 1994; van Gelder, 2002; Biere, Cimatti, Clarke, Fujita, & Zhu, 1999; Stephan, Brayton, & Sangiovanni-Vencentelli, 1996).

The conceptual simplicity of SAT facilitates algorithm development, and considerable research and engineering efforts over the past decades have led to sophisticated algorithms with highly-optimized implementations. In fact, SAT is probably the $\mathcal{NP}$-complete decision problem for which the largest amount of research effort has been expended for the develop-ment and study of algorithms. Today's high-performance SAT solvers include tree-search algorithms (see, e.g., Davis, Logemann, & Loveland, 1962; Zhang, Madigan, Moskewicz, & Malik, 2001; Zhang, 2002; Kullmann, 2002; Dubois & Dequen, 2001; Heule, Zwieten, Dufour, & Maaren, 2004; Eén & Sörensson, 2003), local search algorithms (see, e.g., Selman, Levesque, & Mitchell, 1992; Selman, Kautz, & Cohen, 1994; Hutter, Tompkins, & Hoos, 2002; Hoos, 2002; Li & Huang, 2005; Ishtaiwi, Thornton, Anbulagan, Sattar, & Pham, 2006; Hoos & Stützle, 2005), and resolution-based approaches (see, e.g., Davis & Putnam, 1960; Dechter & Rish, 1994; Bacchus, 2002b, 2002a; Bacchus & Winter, 2003; Subbarayan & Pradhan, 2005).

Most of these SAT algorithms are highly complex, and thus have largely resisted theoretical average-case analysis. Instead, empirical studies are often the only practical means for assessing and comparing their performance. In one prominent and ongoing example, the SAT community holds an annual SAT competition (`http://www.satcompetition.org`; see, e.g., Le Berre & Simon, 2004). This competition is intended to provide an objective assessment of SAT algorithms, and thus to track the state of the art in SAT solving, to assess and promote new solvers, and to identify new challenging benchmarks. Solvers are judged based on their empirical performance on three categories of instances, each of which is further divided into satisfiable, unsatisfiable and mixed instances, with both speed and robustness taken into account. The competition serves as an annual showcase for the state of the art in SAT solving; more than 30 solvers were entered in 2007.

## 1.1 The Algorithm Selection Problem

One way in which evaluations like the SAT competition are useful is that they allow practitioners to determine which algorithm performs best for instances relevant to their problem domain. However, choosing a single algorithm on the basis of competition performance is not always a good approach—indeed, it is often the case that one solver is better than others at solving some problem instances from a given class, but dramatically worse on other instances. Thus, practitioners with hard SAT problems to solve face a potentially difficult "algorithm selection problem" (Rice, 1976): which algorithm(s) should be run in order to minimize some performance objective, such as expected runtime?

The most widely-adopted solution to such algorithm selection problems is to measure every candidate solver's runtime on a representative set of problem instances, and then to use only the algorithm that offered the best (e.g., average or median) performance. We call this the "winner-take-all" approach. Its use has resulted in the neglect of many algorithms that are not competitive on average but that nevertheless offer very good performance on particular instances.

The *ideal* solution to the algorithm selection problem, on the other hand, would be to consult an oracle that knows the amount of time that each algorithm would take to solve a given problem instance, and then to select the algorithm with the best performance. Unfortunately, computationally cheap, perfect oracles of this nature are not available for SAT or any other $\mathcal{NP}$-complete problem; we cannot precisely determine an arbitrary algorithm's runtime on an arbitrary instance without actually running it. Nevertheless, our approach to algorithm selection is based on the idea of building an approximate runtime predictor, which can be seen as a heuristic approximation to a perfect oracle. Specifically, we use machine learning techniques to build an *empirical hardness model*, a computationally inexpensive predictor of an algorithm's runtime on a given problem instance based on features of the instance and the algorithm's past performance (Nudelman, Leyton-Brown, Hoos, Devkar, & Shoham, 2004a; Leyton-Brown, Nudelman, & Shoham, 2002). By modeling several algorithms and, at runtime, choosing the algorithm predicted to have the best performance, empirical hardness models can serve as the basis for an *algorithm portfolio* that solves the algorithm selection problem automatically (Leyton-Brown, Nudelman, Andrew, McFadden, & Shoham, 2003b, 2003a).[1]

In this work we show, for what we believe to be the first time, that empirical hardness models can be used to build an algorithm portfolio that achieves state-of-the-art performance in a broad, practical domain. That is, we evaluated our algorithm not under idiosyncratic conditions and on narrowly-selected data, but rather in a large, independently-conducted competition, confronting a wide range of high-performance algorithms and a large set of independently-chosen interesting data. Specifically, we describe and analyze SATzilla, a portfolio-based SAT solver that utilizes empirical hardness models for per-instance algorithm selection.

## 1.2 Algorithm Portfolios

The term "algorithm portfolio" was introduced by Huberman, Lukose, and Hogg (1997) to describe the strategy of running several algorithms in parallel, potentially with different algorithms being assigned different amounts of CPU time. This approach was also studied by Gomes and Selman (2001). Several authors have since used the term in a broader way that encompasses any strategy that leverages multiple "black-box" algorithms to solve a single problem instance. Under this view, the space of algorithm portfolios is a spectrum, with approaches that use all available algorithms at one end and approaches that always select only a single algorithm at the other. The advantage of using the term *portfolio* to refer to this broader class of algorithms is that they all work for the same reason—they exploit lack of correlation in the best-case performance of several algorithms in order to obtain improved performance in the average case.

To more clearly describe algorithm portfolios in this broad sense, we introduce some new terminology. We define an $(a, b)$-of-$n$ portfolio as a set of $n$ algorithms and a procedure for selecting among them with the property that if no algorithm terminates early, at least $a$ and no more than $b$ algorithms will be executed.[2] For brevity, we also use the terms $a$-of-$n$ portfolio to refer to an $(a, a)$-of-$n$ portfolio, and $n$-portfolio for an $n$-of-$n$ portfolio. It is also useful to distinguish how solvers are run after being selected. Portfolios can be *parallel* (all algorithms are executed concurrently), *sequential* (the execution of one algorithm only begins when the previous algorithm's execution has ended), or *partly sequential* (some combination of the two). Thus the "classic" algorithm portfolios of Huberman et al. (1997) and Gomes and Selman (2001) can be described as parallel $n$-portfolios. In contrast, the `SATzilla` solvers that we will present in this paper are sequential 3-of-$n$ portfolios since they sequentially execute two pre-solvers followed by one main solver.

---

1. Similarly, one could predict the performance of a single algorithm under different parameter settings and choose the best setting on a per-instance basis. We have previously demonstrated that this approach is feasible in the case where the number of parameters is small (Hutter, Hamadi, Hoos, & Leyton-Brown, 2006). Ultimately, it is conceivable to combine the two lines of research, and to automatically select a good algorithm along with good parameter settings on a per-instance basis.

2. The termination condition is somewhat tricky. We consider the portfolio to have terminated early if it solves the problem before one of the solvers has a chance to run, or if one of the solvers crashes. Thus, when determining $a$ and $b$, we do not consider crash-recovery techniques, such as using the next best predicted solver (discussed later in this paper).

There is a range of other work in the literature that describes algorithm portfolios in the broad sense that we have defined here. First, we consider work that has emphasized algorithm selection (or 1-of-$n$ portfolios). Lobjois and Lemaître (1998) studied the problem of selecting between branch-and-bound algorithms based on an estimate of search tree size due to Knuth (1975). Gebruers, Hnich, Bridge, and Freuder (2005) employed case-based reasoning to select a solution strategy for instances of a constraint programming problem. Various authors have proposed classification-based methods for algorithm selection (e.g., Guerri & Milano, 2004; Gebruers, Guerri, Hnich, & Milano, 2004; Guo & Hsu, 2004; and, to some extent, Horvitz, Ruan, Gomes, Kautz, Selman, & Chickering, 2001). One problem with such approaches is that they use an error metric that penalizes all misclassifications equally, regardless of their cost. This is problematic because using a suboptimal algorithm is acceptable, provided it is *nearly* as good as the best algorithm. Our `SATzilla` approach can be considered to be a classifier with an error metric that depends on the difference in runtime between algorithms.

At the other end of the spectrum, much work has been done that considers switching between multiple algorithms, or in our terminology building parallel $n$-portfolios. Gomes and Selman (2001) built a portfolio of stochastic algorithms for quasi-group completion and logistics scheduling problems. Low-knowledge algorithm control by Carchrae and Beck (2005) employed a portfolio of anytime algorithms, prioritizing each algorithm according to its performance so far. Gagliolo and Schmidhuber (2006b) learned dynamic algorithm portfolios that also support running several algorithms at once, where an algorithm's priority depends on its predicted runtime conditioned on the fact that it has not yet found a solution. Streeter, Golovin, and Smith (2007) improved average-case performance by using black-box techniques for learning how to interleave the execution of multiple heuristics based not on instance features but only on the runtime of algorithms.

Some approaches fall between these two extremes, making decisions about which algorithms to use on the fly—while solving a problem instance—instead of committing in advance to a subset of algorithms. The examples we give here are $(1, n)$-of-$n$ portfolios. Lagoudakis and Littman (2001) employed reinforcement learning to solve an algorithm selection problem at each decision point of a DPLL solver for SAT in order to select a branching rule. Similarly, Samulowitz and Memisevic (2007) employed classification to switch between different heuristics for QBF solving during the search.

Finally, we describe the ways in which this paper builds on our own past work. Leyton-Brown et al. (2002) introduced empirical hardness models. Nudelman et al. (2004a) demonstrated that they work on (uniform-random) SAT and introduced the features that we use here, and Hutter et al. (2006) showed how to apply them to randomized, incomplete algorithms. Empirical hardness models were first used as a basis for algorithm portfolios by Leyton-Brown et al. (2003b, 2003a). The idea of building such an algorithm portfolio for SAT goes back to 2003, when we submitted the first `SATzilla` solver to the SAT competition (Nudelman, Leyton-Brown, Devkar, Shoham, & Hoos, 2004b); this version of `SATzilla` placed 2[nd] in two categories and 3[rd] in another. In the following, we describe a substantially improved `SATzilla` solver, which was entered into the 2007 SAT Competition and—despite considerable progress in the SAT community over this four year interval—placed 1[st] in three categories, and 2[nd] and 3[rd] in two further categories. This solver was described, along with some preliminary analysis, in a conference paper (Xu, Hutter, Hoos, & Leyton-Brown, 2007c); it used *hierarchical hardness models*, described separately (Xu, Hoos, & Leyton-Brown, 2007a). In this work, we provide a much more detailed description of this new solver, present several new techniques that have never been previously published (chiefly introduced in Section 5) and report new experimental results.

## 1.3 Overview

Overall, this paper is divided into two parts. The first part describes the development of `SATzilla07`, which we submitted to the 2007 SAT Competition and the second part demonstrates our most recent, improved portfolio algorithms (`SATzilla07`[+] and `SATzilla07`[*]). Each part is subdivided into three sections, the first of which describes our approach for designing a portfolio-based solver at a high level, the second of which explains lower-level details of portfolio construction, and the third of which provides the results of an extensive experimental evaluation.

Section 2 (Design I) begins with a general methodology for building algorithm portfolios based on empirical hardness models. In this work, we apply these general strategies to SAT and evaluate them experimentally. In Section 3 (Construction I) we describe the architecture of the portfolio-based solvers that we entered into the 2007 SAT Competition and described in our previous work (Xu et al., 2007c). In addition, we constructed a new portfolio-based solver for `INDUSTRIAL` instances and analyzed it by effectively re-running the `INDUSTRIAL` category of the 2007 competition with our portfolio included; the results of this analysis are reported in Section 4 (Evaluation I).

We then move on to consider ways of extending, strengthening, and automating our portfolio construction. We present five new design ideas in Section 5 (Design II), and consider the incorporation of new solvers and training data in Section 6 (Construction II). Finally, we evaluated these new ideas and quantified their benefits in a second set of experiments, which we describe in Section 7 (Evaluation II). Section 8 concludes the paper with some general observations.

## 2. Design I: Building Algorithm Portfolios with Empirical Hardness Models

The general methodology for building an algorithm portfolio that we use in this work follows Leyton-Brown et al. (2003b) in its broad strokes, but we have made significant extensions here. Portfolio construction happens offline, as part of algorithm development, and comprises the following steps.

1. Identify a target distribution of problem instances. Practically, this means selecting a set of instances believed to be representative of some underlying distribution, or using an instance generator that constructs instances that represent samples from such a distribution.

2. Select a set of candidate solvers that have relatively uncorrelated runtimes on this distribution and are known or expected to perform well on at least some instances.

3. Identify features that characterize problem instances. In general this cannot be done automatically, but rather must reflect the knowledge of a domain expert. To be usable effectively for automated algorithm selection, these features must be related to instance hardness and relatively cheap to compute.

4. On a training set of problem instances, compute these features and run each algorithm to determine its running times.

5. Identify one or more solvers to use for pre-solving instances. These pre-solvers will later be run for a short amount of time before features are computed (step 9 below), in order to ensure good performance on very easy instances and to allow the empirical hardness models to focus exclusively on harder instances.

6. Using a validation data set, determine which solver achieves the best performance for all instances that are not solved by the pre-solvers and on which the feature computation times out. We refer to this solver as the *backup solver*. In the absence of a sufficient number of instances for which pre-solving and feature computation timed out, we employ the single best component solver (i.e., the winner-take-all choice) as a backup solver.

7. Construct an empirical hardness model for each algorithm in the portfolio, which predicts the runtime of the algorithm for each instance, based on the instance's features.

8. Choose the best subset of solvers to use in the final portfolio. We formalize and automatically solve this as a simple subset selection problem: from all given solvers, select a subset for which the respective portfolio (which uses the empirical hardness models learned in the previous step) achieves the best performance on the validation set. (Observe that because our runtime predictions are not perfect, dropping a solver from the portfolio entirely can increase the portfolio's overall performance.)

Then, online, to solve a given problem instance, the following steps are performed.

9. Run each pre-solver until a predetermined fixed cutoff time is reached.

10. Compute feature values. If feature computation cannot be completed for some reason (error or timeout), select the backup solver identified in step 6 above.

11. Otherwise, predict each algorithm's runtime using the empirical hardness models from step 7 above.

12. Run the algorithm predicted to be the best. If a solver fails to complete its run (e.g., it crashes), run the algorithm predicted to be next best.

The effectiveness of an algorithm portfolio built using this approach depends on our ability to learn empirical hardness models that can accurately predict a solver's runtime on a given instance using efficiently computable features. In the experiments presented in this paper, we use the same ridge regression method that has previously proven to be very successful in predicting runtime on uniform random $k$-SAT, on structured SAT instances, and on combinatorial auction winner determination problems (Nudelman et al., 2004a; Hutter et al., 2006; Leyton-Brown et al., 2002).[3]

## 2.1 Ridge Regression and Feature Selection

We now explain the construction of the empirical hardness models described in Step 7 above. To predict the runtime of an algorithm $\mathcal{A}$ on an instance distribution $\mathcal{D}$, we first draw $n$ instances from $\mathcal{D}$ uniformly at random. (In this article, the distributions are given implicitly by a benchmark set of instances, and we simply use all instances in the benchmark set.) For each instance $i$, we compute a set of features $\boldsymbol{x}_i = [x_{i,1}, \ldots, x_{i,m}]$ that characterize the instance. We also run algorithm $\mathcal{A}$ on the instance, recording its runtime $r_i$.

Having computed features and runtimes on all $n$ instances, we fit a function $f(\boldsymbol{x})$ that, given the features $\boldsymbol{x}_i$ of instance $i$, yields a prediction, $\tilde{y}_i$, of the logarithm of $\mathcal{A}$'s runtime $y_i = \log r_i$. In our experience, we have found this log transformation of runtime to be very important due to the large variation in runtimes for hard combinatorial problems. Unfortunately, the performance of learning algorithms can deteriorate when some features are uninformative or highly correlated with other features; it is difficult to construct features that do not suffer from these problems. Therefore, we first reduce the set of features by performing feature selection, in our case forward selection (see e.g., Guyon, Gunn, Nikravesh, & Zadeh, 2006), a simple iterative method that starts with an empty feature set and greedily adds one feature at a time, aiming to reduce cross-validation error as much as possible with every added feature. Next, we add additional pairwise product features $x_{i,j} \cdot x_{i,k}$ for $j = 1 \ldots m$ and $k = j \ldots m$; this is a widely used method typically referred to as quadratic basis function expansion. Finally, we perform another pass of forward selection on this extended set to determine our final set of basis functions, such that for instance $i$ we obtain an expanded feature vector $\boldsymbol{\phi}_i = \boldsymbol{\phi}(\boldsymbol{x}_i) = [\phi_1(\boldsymbol{x}_i), \ldots, \phi_d(\boldsymbol{x}_i)]$, where $d$ is the final number of basis functions used in the model.

We then use ridge regression (see, e.g., Bishop, 2006) to fit the free parameters $\boldsymbol{w}$ of the function $f_{\boldsymbol{w}}(\boldsymbol{x})$. Ridge regression works as follows. Let $\boldsymbol{\Phi}$ be an $n \times d$ matrix containing the vectors $\boldsymbol{\phi}_i$ for each instance in the training set, let $\boldsymbol{y}$ be the vector of log runtimes, and let $I$ be the $d \times d$ identity matrix. Finally, let $\delta$ be a small constant to penalize large coefficients $\boldsymbol{w}$ and thereby increase numerical stability (we used $\delta = 10^{-3}$ in our experiments). Then, we compute $\boldsymbol{w} = (\delta I + \boldsymbol{\Phi}^\top \boldsymbol{\Phi})^{-1} \boldsymbol{\Phi}^\top \boldsymbol{y}$, where $\boldsymbol{\Phi}^\top$ denotes the transpose of matrix $\boldsymbol{\Phi}$. For a previously unseen instance $i$, we obtain a prediction of log runtime by computing the instance features $\boldsymbol{x}_i$ and evaluating $f_{\boldsymbol{w}}(\boldsymbol{x}_i) = \boldsymbol{w}^\top \boldsymbol{\phi}(\boldsymbol{x}_i)$.

---

3. It should be noted that our portfolio methodology can make use of any regression approach that provides sufficiently accurate estimates of an algorithm's runtime and that is computationally efficient enough that the time spent making a prediction can be compensated for by the performance gain obtained through improved algorithm selection. For example, in similar settings, we have previously explored many other learning techniques, such as lasso regression, SVM regression, and Gaussian process regression (Leyton-Brown et al., 2002; Hutter et al., 2006). All of these techniques are computationally more expensive than ridge regression, and in our previous experiments we found that they did not improve predictive performance enough to justify this additional cost.

## 2.2 Accounting for Censored Data

As is common with heuristic algorithms for solving $\mathcal{NP}$-complete problems, SAT algorithms tend to solve some instances very quickly, while taking an extremely long amount of time to solve other instances. Hence, runtime data can be very costly to gather, as individual runs can take literally weeks to complete, even when other runs on instances of the same size take only milliseconds. The common solution to this problem is to "censor" some runs by terminating them after a fixed cutoff time.

The question of how to fit good models in the presence of censored data has been extensively studied in the survival analysis literature in statistics, which originated in actuarial questions such as estimating a person's lifespan given mortality data as well as the ages and characteristics of other people still alive. Observe that this problem is the same as ours, except that in our case, data points are always censored at the same value, a subtlety that turns out not to matter.

The best approach that we know for dealing with censored data is to build models that use all available information about censored runs by using the censored runtimes as lower bounds on the actual runtimes. To our knowledge, this technique was first used in the context of SAT by Gagliolo and Schmidhuber (2006a). We chose the simple, yet effective method by Schmee and Hahn (1979) to deal with censored samples. In brief, this method first trains a hardness model treating the cutoff time as the true (uncensored) runtime for censored samples, and then repeats the following steps until convergence.

1. Estimate the expected runtime of censored runs using the hardness model. Since in ridge regression, predictions are in fact normal distributions (with a fixed variance), the expected runtime conditional on the runtime exceeding the cutoff time is the mean of the corresponding normal distribution truncated at the cutoff time.

2. Train a new hardness model using true runtimes for the uncensored instances and the predictions generated in the previous step for the censored instances.

In earlier work (Xu, Hutter, Hoos, & Leyton-Brown, 2007b), we experimentally compared this approach with two other approaches for dealing with censored data: dropping such data entirely, and treating censored runs as though they had finished at the cutoff threshold. We demonstrated empirically that both of these methods are significantly worse than the method presented above. Intuitively, both methods introduce bias into empirical hardness models, whereas the method by Schmee and Hahn (1979) is unbiased.

## 2.3 Using Hierarchical Hardness Models

Our previous research on empirical hardness models for SAT showed that we can achieve better prediction accuracy even with simpler models if we restrict ourselves only to satisfiable or unsatisfiable instances (Nudelman et al., 2004a). Of course, in practice we are interested in making accurate predictions even when we do not know whether an instance is satisfiable. In recent work (Xu et al., 2007a), we introduced *hierarchical hardness models* as a method for solving this problem. We define the subjective probability that an instance with features $x$ is satisfiable to be the probability that an instance chosen at random from the underlying instance distribution with features matching $x$ is satisfiable. Hierarchical hardness models first use a classifier to predict this subjective probability of satisfiability and then use this probability, as well as the features $\boldsymbol{x}$, to combine the predictions of so-called *conditional models*, which are trained only on satisfiable instances and only on unsatisfiable instances, respectively. In our previous work we conducted extensive experiments on various types of SAT instances and found that these hierarchical hardness models achieve better runtime prediction accuracies than traditional empirical hardness models (Xu et al., 2007a).

Specifically, we begin by predicting an instance's satisfiability using a classification algorithm that depends on the same instance features used by the empirical hardness models. We chose Sparse Multinomial Logistic Regression, SMLR (Krishnapuram, Carin, Figueiredo, & Hartemink, 2005), but any other classification algorithm that returns the probability that an instance belongs to each class could be used instead. Then, we train conditional empirical hardness models ($M_{sat}$, $M_{unsat}$) using quadratic basis-function regression for both satisfiable and unsatisfiable training instances.

Next, we must decide how to combine the predictions of these two models.[4] We observe a set of instance features $\boldsymbol{x}$ and a classifier prediction $s$; our task is to predict the expected value of the algorithm's runtime $y$ given this information. We introduce an additional random variable $z \in \{sat, unsat\}$, which represents our subjective belief about an oracle's choice of which conditional model will perform best for a given instance. (Observe that this may not always correspond to the model trained on the data with the same satisfiability status as the instance.) We can express the conditional dependence relationships between our random variables using a graphical model, as illustrated in Figure 1.

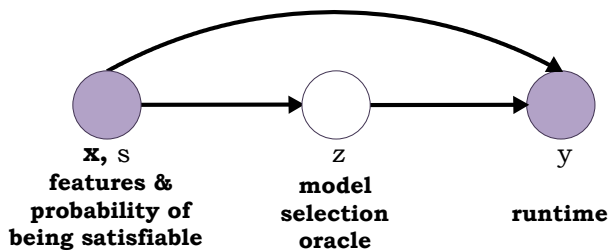

Figure 1: *Graphical model for our mixture-of-experts approach.*

Then we can write the expression for the probability distribution over an instance's runtime given the features $\boldsymbol{x}$ and $s$ as

$$P(y \mid \boldsymbol{x}, s) = \sum_{k \in \{sat, unsat\}} P(z = k \mid \boldsymbol{x}, s) \cdot P_{M_k}(y \mid \boldsymbol{x}, s), \qquad (1)$$

where $P_{M_k}(y \mid \boldsymbol{x}, s)$ is the probability of $y$ evaluated according to model $M_k$ (see Figure 1). Since the models were fitted using ridge regression, we can rewrite Equation (1) as

$$P(y \mid \boldsymbol{x}, s) = \sum_{k \in \{sat, unsat\}} P(z = k \mid \boldsymbol{x}, s) \cdot \varphi \left( \frac{y - \boldsymbol{w}_k^\top \phi(\boldsymbol{x})}{\sigma_k} \right), \qquad (2)$$

where $\varphi(\cdot)$ denotes the probability distribution function of a Normal distribution with mean zero and unit variance, $\boldsymbol{w}_k$ are the weights of model $M_k$, $\phi(\boldsymbol{x})$ is the quadratic basis function expansion of $\boldsymbol{x}$, and $\sigma_k$ is a fixed standard deviation.

---

4. Note that the classifier's output is not used directly to select a model—doing so would mean ignoring the cost of making a mistake. Instead, we use the classifier's output as a feature upon which our hierarchical model can depend.

In particular, we are interested in the mean predicted runtime, that is, the expectation of $P(y \mid \boldsymbol{x}, s)$:

$$\mathbb{E}(y \mid \boldsymbol{x}, s) = \sum_{k \in \{sat, unsat\}} P(z = k \mid \boldsymbol{x}, s) \cdot \boldsymbol{w}_k^\top \phi(\boldsymbol{x}). \qquad (3)$$

Evaluating this expression would be easy if we knew $P(z = k \mid \boldsymbol{x}, s)$; of course, we do not. Our approach is to learn weighting functions $P(z = k \mid \boldsymbol{x}, s)$ to minimize the following loss function:

$$\mathcal{L} = \sum_{i=1}^{n} \left( y_i - \mathbb{E}(y \mid \boldsymbol{x}, s) \right)^2, \qquad (4)$$

where $y_i$ is the true log runtime on instance $i$ and $n$ is the number of training instances.

As the hypothesis space for these weighting functions we chose the softmax function (see, e.g., Bishop, 2006)

$$P(z = sat \mid \boldsymbol{x}, s) = \frac{e^{\boldsymbol{v}^\top [\boldsymbol{x}; s]}}{1 + e^{\boldsymbol{v}^\top [\boldsymbol{x}; s]}}, \qquad (5)$$

where $\boldsymbol{v}$ is a vector of free parameters that is set to minimize the loss function (4). This functional form is frequently used for probabilistic classification tasks: if $\boldsymbol{v}^\top [\boldsymbol{x}; s]$ is large and positive, then $e^{\boldsymbol{v}^\top [\boldsymbol{x}; s]}$ is much larger than 1 and the resulting probability is close to 1; if it is large and negative, the result is close to zero; if it is zero, then the resulting probability is exactly 0.5.

This can be seen as a mixture-of-experts problem (see, e.g., Bishop, 2006) with the experts fixed to $M_{sat}$ and $M_{unsat}$. (In traditional mixture-of-experts methods, the experts are allowed to vary during the training process described below.) For implementation convenience, we used an existing mixture of experts implementation to optimize $\boldsymbol{v}$, which is built around an expectation maximization (EM) algorithm that performs iterative reweighted least squares in the M step (Murphy, 2001). We modified this code slightly to fix the experts and initialized the choice of expert to the classifier's output by setting the initial values of $P(z \mid \boldsymbol{x}, s)$ to $s$. Note that other numerical optimization procedures could be used to minimize the loss function (4) with respect to $\boldsymbol{v}$.

Having optimized $\boldsymbol{v}$, to obtain a runtime prediction for an unseen test instance we simply compute the instance's features $\boldsymbol{x}$ and the classifier's output $s$, and then compute the expected runtime by evaluating Equation (3).

Finally, note that these techniques do not require us to restrict ourselves to the conditional models $M_{sat}$ and $M_{unsat}$, or even to the use of only two models. In Section 7, we describe hierarchical hardness models that rely on *six* conditional models, trained on satisfiable and unsatisfiable instances from different data sets.

## 3. Construction I: Building `SATzilla07` for the 2007 SAT Competition

In this section, we describe the `SATzilla07` solvers entered into the 2007 SAT Competition, which—like previous events in the series—featured three main categories of instances, `RANDOM`, `HANDMADE` (also known as `CRAFTED`) and `INDUSTRIAL`. We submitted three versions of `SATzilla07` to the competition. Two versions specifically targeted the `RANDOM` and `HANDMADE` instance categories and were trained only on data from their target category. In order to allow us to study `SATzilla07`'s performance on an even more heterogeneous instance distribution, a third version of `SATzilla07` was trained on data from all three categories of the competition; we call this new category `ALL`. We did not construct a version of `SATzilla07` for the `INDUSTRIAL` category, because of time constraints and the limit of three submissions per team. However, we built such a version after the submission deadline and report results for it below.

All of our solvers were built using the design methodology detailed in Section 2. Each of the following subsections corresponds to one step from this methodology.

### 3.1 Selecting Instances

In order to train empirical hardness models for any of the above scenarios, we needed instances that would be similar to those used in the real competition. For this purpose we used instances from the respective categories of all previous SAT competitions (2002, 2003, 2004, and 2005), as well as from the 2006 SAT Race (which only featured `INDUSTRIAL` instances). Instances that were repeated in previous competitions were also repeated in our data sets. Overall, there were 4 811 instances: 2 300 instances in category `RANDOM`, 1 490 in category `HANDMADE` and 1 021 in category `INDUSTRIAL`; of course, category `ALL` included all of these instances. About 68% of the instances were solved within 1 200 CPU seconds on our reference machine by at least one of the seven solvers we used (see Section 3.2 below; the computational infrastructure used for our experiments is described in Section 3.4). All instances that were not solved by any of these solvers were dropped from our data set.

We randomly split our data set into training, validation and test sets at a ratio of 40:30:30. All parameter tuning and intermediate model testing was performed on the validation set; the test set was used only to generate the final results reported here.

In this section, we use the same `SATzilla07` methodology for building multiple portfolios for different sets of benchmark instances. In order to avoid confusion between the changes to our overall methodology discussed later and differences in training data, we treat the data set as an input parameter for `SATzilla`. For the data set comprising the previously mentioned `RANDOM` instances, we write $D_r$; similarly, we write $D_h$ for `HANDMADE` and $D_i$ for `INDUSTRIAL`; for `ALL`, we simply write $D$.

### 3.2 Selecting Solvers

To decide what algorithms to include in our portfolio, we considered a wide variety of solvers that had been entered into previous SAT competitions and into the 2006 SAT Race. We manually analyzed the results of these competitions, identifying all algorithms that yielded the best performance on some subset of instances. Since our focus was on both satisfiable and unsatisfiable instances, and since we were concerned about the cost of misclassifications, we did not choose any incomplete algorithms at this stage; however, we revisit this issue in Section 5.4. In the end, we selected the seven high-performance solvers shown in Table 1 as candidates for the `SATzilla07` portfolio. Like the data set used for training, we treat the component solvers as an input to `SATzilla`, and denote the set of solvers from Table 1 as `S`.

| Solver | Reference |
|---|---|
| Eureka | Nadel, Gordon, Palti, and Hanna (2006) |
| Kcnfs06 | Dubois and Dequen (2001) |
| March_dl04 | Heule et al. (2004) |
| Minisat 2.0 | Eén and Sörensson (2006) |
| Rsat 1.03 | Pipatsrisawat and Darwiche (2006) |
| Vallst | Vallstrom (2005) |
| Zchaff_Rand | Mahajan, Fu, and Malik (2005) |

Table 1: The seven solvers in `SATzilla07`; we refer to this set of solvers as `S`.

In previous work (Xu et al., 2007b), we considered using the Hypre preprocessor (Bacchus & Winter, 2003) before applying one of `SATzilla`'s component solvers; this effectively doubled our number of component solvers. For this work, we re-evaluated this option and found performance to basically remain unchanged without preprocessing (performance differences in terms of instances solved, runtime, and SAT competition score were smaller than 1%, and even this small difference was not consistently in favor of using the Hypre preprocessor). For this reason, we dropped preprocessing in the work reported here.

## 3.3 Choosing Features

The choice of instance features has a significant impact on the performance of empirical hardness models. Good features need to correlate well with (solver-specific) instance hardness and need to be cheap to compute, since feature computation time counts as part of `SATzilla07`'s runtime.

Nudelman et al. (2004a) introduced 84 features for SAT instances. These features can be classified into nine categories: problem size, variable-clause graph, variable graph, clause graph, balance, proximity to Horn formulae, LP-based, DPLL probing and local search probing features — the code for this last group of features was based on `UBCSAT` (Tompkins & Hoos, 2004). In order to limit the time spent computing features, we slightly modified the feature computation code of Nudelman et al. (2004a). For `SATzilla07`, we excluded a number of computationally expensive features, such as LP-based and clause graph features. The computation time for each of the local search and DPLL probing features was limited to 1 CPU second, and the total feature computation time per instance was limited to 60 CPU seconds. After eliminating some features that had the same value across all instances and some that were too unstable given only 1 CPU second of local search probing, we ended up using the 48 raw features summarized in Figure 2.

**Problem Size Features:**
1. **Number of clauses**: denoted c
2. **Number of variables**: denoted v
3. **Ratio**: c/v

**Variable-Clause Graph Features:**
4-8. **Variable nodes degree statistics:** mean, variation coefficient, min, max and entropy.
9-13. **Clause nodes degree statistics:** mean, variation coefficient, min, max and entropy.

**Variable Graph Features:**
14-17. **Nodes degree statistics:** mean, variation coefficient, min and max.

**Balance Features:**
18-20. **Ratio of positive and negative literals in each clause:** mean, variation coefficient and entropy.
21-25. **Ratio of positive and negative occurrences of each variable:** mean, variation coefficient, min, max and entropy.
26-27. **Fraction of binary and ternary clauses**

**Proximity to Horn Formula:**
28. **Fraction of Horn clauses**
29-33. **Number of occurrences in a Horn clause for each variable:** mean, variation coefficient, min, max and entropy.

**DPLL Probing Features:**
34-38. **Number of unit propagations:** computed at depths 1, 4, 16, 64 and 256.
39-40. **Search space size estimate:** mean depth to contradiction, estimate of the log of number of nodes.

**Local Search Probing Features:**
41-44. **Number of steps to the best local minimum in a run:** mean, median, 10th and 90th percentiles for SAPS.
45. **Average improvement to best in a run:** mean improvement per step to best solution for SAPS.
46-47. **Fraction of improvement due to first local minimum:** mean for SAPS and GSAT.
48. **Coefficient of variation of the number of unsatisfied clauses in each local minimum:** mean over all runs for SAPS.

Figure 2: *The features used for building* `SATzilla07`*; these were originally introduced and described in detail by Nudelman et al. (2004a).*

## 3.4 Computing Features and Runtimes

All our experiments were performed using a computer cluster consisting of 55 machines with dual Intel Xeon 3.2GHz CPUs, 2MB cache and 2GB RAM, running Suse Linux 10.1. As in the SAT competition, all runs of any solver that exceeded a certain runtime were aborted (censored) and recorded as such. In order to keep the computational cost manageable, we chose a cutoff time of 1 200 CPU seconds.

## 3.5 Identifying Pre-solvers

As described in Section 2, in order to solve easy instances quickly without spending any time for the computation of features, we use one or more *pre-solvers*: algorithms that are run unconditionally but briefly before features are computed. Good algorithms for pre-solving solve a large proportion of instances quickly. Based on an examination of the training runtime data, we chose `March_dl04` and the local search algorithm `SAPS` (Hutter et al., 2002) as pre-solvers for `RANDOM`, `HANDMADE` and `ALL`; for `SAPS`, we used the `UBCSAT` implementation (Tompkins & Hoos, 2004) with the best fixed parameter configuration identified by Hutter et al. (2006). (Note that while we did not consider incomplete algorithms for inclusion in the portfolio, we did use one here.)

Within 5 CPU seconds on our reference machine, `March_dl04` solved 47.8%, 47.7%, and 43.4% of the instances in our `RANDOM`, `HANDMADE` and `ALL` data sets, respectively. For the remaining instances, we let `SAPS` run for 2 CPU seconds, because we found its runtime to be almost completely uncorrelated with `March_dl04` (Pearson correlation coefficient $r = 0.118$ for the 487 remaining instances solved by both solvers). In this time, `SAPS` solved 28.8%, 5.3%, and 14.5% of the remaining `RANDOM`, `HANDMADE` and `ALL` instances, respectively. For the `INDUSTRIAL` category, we chose to run `Rsat 1.03` for 2 CPU seconds as a pre-solver, which resulted in 32.0% of the instances in our `INDUSTRIAL` set being solved. Since `SAPS` solved less than 3% of the remaining instances within 2 CPU seconds, it was not used as a pre-solver in this category.

| Solver | RANDOM | | HANDMADE | | INDUSTRIAL | | ALL | |
|---|---|---|---|---|---|---|---|---|
| | Time | Solved | Time | Solved | Time | Solved | Time | Solved |
| Eureka | 770 | 40% | 561 | 59% | **330** | **84%** | 598 | 57% |
| Kcnfs06 | 319 | 81% | 846 | 33% | 1050 | 13% | 658 | 50% |
| March_dl04 | **269** | **85%** | **311** | **80%** | 715 | 42% | **394** | **73%** |
| Minisat 2.0 | 520 | 62% | 411 | 73% | 407 | 76% | 459 | 69% |
| Rsat 1.03 | 522 | 62% | 412 | 72% | 345 | 81% | 445 | 70% |
| Vallst | 757 | 40% | 440 | 67% | 582 | 59% | 620 | 54% |
| Zchaff_Rand | 802 | 36% | 562 | 58% | 461 | 71% | 645 | 51% |

Table 2: *Average runtime (in CPU seconds on our reference machine) and percentage of instances solved by each solver for all instances that were solved by at least one of our seven component solvers within the cutoff time of 1 200 seconds.*

### 3.6 Identifying the Backup Solver

The performance of all our solvers from Table 1 is reported in Table 2. We computed average runtime (here and in the remainder of this work) counting timeouts as runs that completed at the cutoff time of 1 200 CPU seconds. As can be seen from this data, the best single solver for `ALL`, `RANDOM` and `HANDMADE` was always `March_dl04`. For categories `RANDOM` and `HANDMADE`, we did not encounter instances for which the feature computation timed out. Thus, we employed the winner-take-all solver `March_dl04` as a backup solver in both of these domains. For categories `INDUSTRIAL` and `ALL`, `Eureka` performed best on those instances that remained unsolved after pre-solving and for which feature computation timed out; we thus chose `Eureka` as the backup solver.

### 3.7 Learning Empirical Hardness Models

We learned empirical hardness models for predicting each solver's runtime as described in Section 2, using the procedure of Schmee and Hahn (1979) for dealing with censored data and also employing hierarchical hardness models.

### 3.8 Solver Subset Selection

We performed automatic exhaustive subset search as outlined in Section 2 to determine which solvers to include in `SATzilla07`. Table 3 describes the solvers that were selected for each of our four data sets.

| Data Set | Solvers used in `SATzilla07` |
|---|---|
| RANDOM | `March_dl04, Kcnfs06, Rsat 1.03` |
| HANDMADE | `Kcnfs06, March_dl04, Minisat 2.0, Rsat 1.03, Zchaff_Rand` |
| INDUSTRIAL | `Eureka, March_dl04, Minisat 2.0, Rsat 1.03` |
| ALL | `Eureka, Kcnfs06, March_dl04, Minisat 2.0, Zchaff_Rand` |

Table 3: Results of subset selection for `SATzilla07`.

## 4. Evaluation I: Performance Analysis of `SATzilla07`

In this section, we evaluate `SATzilla07` for our four data sets. Since we use the SAT Competition as a running example throughout this paper, we start by describing how `SATzilla07` fared in the 2007 SAT Competition. We then describe more comprehensive evaluations of each `SATzilla07` version in which we compared it in greater detail against its component solvers.

### 4.1 `SATzilla07` in the 2007 SAT Competition

We submitted three versions of `SATzilla07` to the 2007 SAT Competition, namely `SATzilla07(S,D_r)` (i.e., `SATzilla07` using the seven component solvers from Table 1 and trained on RANDOM instances) `SATzilla07(S,D_h)` (trained on HANDMADE), and `SATzilla07(S,D)` (trained on ALL). Table 4 shows the results of the 2007 SAT Competition for the RANDOM and HANDMADE categories. In the RANDOM category, `SATzilla07(S,D_r)` won the gold medal in the subcategory SAT+UNSAT, and came third in the UNSAT subcategory. The SAT subcategory was dominated by local search solvers. In the HANDMADE category, `SATzilla07(S,D_h)` showed excellent performance, winning the SAT+UNSAT and UNSAT subcategories, and placing second in the SAT subcategory.

| Category | Rank | SAT & UNSAT | SAT | UNSAT |
|---|---|---|---|---|
| RANDOM | 1$^{st}$ | `SATzilla07(S,D`$_r$`)` | `Gnovelty+` | `March_ks` |
| | 2$^{nd}$ | `March_ks` | `Ag2wsat0` | `Kcnfs04` |
| | 3$^{rd}$ | `Kcnfs04` | `Ag2wsat+` | `SATzilla07(S,D`$_r$`)` |
| HANDMADE | 1$^{st}$ | `SATzilla07(S,D`$_h$`)` | `March_ks` | `SATzilla07(S,D`$_h$`)` |
| | 2$^{nd}$ | `Minisat07` | `SATzilla07(S,D`$_h$`)` | `TTS` |
| | 3$^{rd}$ | `MXC` | `Minisat07` | `Minisat07` |

Table 4: Results from the 2007 SAT Competition. More that 30 solvers competed in each category.

Since the general portfolio `SATzilla07(S,D)` included `Eureka` (whose source code is not publicly available) it was run in the demonstration division only. The official competition results (available at `http://www.cril.univ-artois.fr/SAT07/`) show that this solver, which was trained on instances from all three categories, performed very well, solving more instances of the union of the three categories than any other solver (including the other two versions of `SATzilla07`).

| Solver | Average runtime [s] | Solved percentage | Performance score |
|--------|--------------------:|:-----------------:|:-----------------:|
| `Picosat` | 398 | 82 | 31484 |
| `TinisatElite` | 494 | 71 | 21630 |
| `Minisat07` | 484 | 72 | **34088** |
| `Rsat 2.0` | 446 | 75 | 23446 |
| `SATzilla07(S,D`$_i$`)` | **346** | **87** | 29552 |

Table 5: *Performance comparison of* `SATzilla07(S,D`$_i$`)` *and winners of the 2007 SAT Competition in the* `INDUSTRIAL` *category. The performance scores are computed using the 2007 SAT Competition scoring function with a cutoff time of* 1 200 *CPU seconds.* `SATzilla07(S,D`$_i$`)` *is exactly the same solver as shown in Figure 6 and was trained without reference to the 2007 SAT Competition data.*

We did not submit a version of `SATzilla07` to the 2007 SAT Competition that was specifically trained on instances from the `INDUSTRIAL` category. However, we constructed such a version, `SATzilla07(S,D`$_i$`)`, after the submission deadline and here report on its performance. Although `SATzilla07(S,D`$_i$`)` did not compete in the actual 2007 SAT Competition, we can approximate how well it would have performed in a simulation of the competition, using the same scoring function as in the competition (described in detail in Section 5.3), based on a large number of competitors, namely the 19 solvers listed in Tables 1, 9 and 10, plus `SATzilla07(S,D`$_i$`)`.

Table 5 compares the performance of `SATzilla07` in this simulation of the competition against all solvers that won at least one medal in the `INDUSTRIAL` category of the 2007 SAT Competition: `Picosat`, `TinisatElite`, `Minisat07` and `Rsat 2.0`. There are some differences between our test environment and that used in the real SAT competition: our simulation ran on different machines and under a different operating system; it also used a shorter cutoff time and fewer competitors to evaluate the solver's performance scores. Because of these differences, the ranking of solvers in our simulation is not necessarily the same as it would have been in the actual competition. Nevertheless, our results leave no doubt that `SATzilla07` can compete with the state-of-the-art SAT solvers in the `INDUSTRIAL` category.

## 4.2 Feature Computation

The actual time required to compute our features varied from instance to instance. In the following, we report runtimes for computing features for the instance sets defined in Section 3.1. Typically, feature computation took at least 3 CPU seconds: 1 second each for local search probing with `SAPS` and `GSAT`, and 1 further second for DPLL probing. However, for some small instances, the limit of 300 000 local search steps was reached before one CPU second had passed, resulting in feature computation times lower than 3 CPU seconds. For most instances from the `RANDOM` and `HANDMADE` categories, the computation of the other features took an insignificant amount of time, resulting in feature computation times just above 3 CPU seconds. However, for many instances from the INDUSTRIAL category, the feature computation was quite expensive, with times up to 1 200 CPU seconds for some instances. We limited the feature computation time to 60 CPU seconds, which resulted in time-outs for 19% of the instances from the INDUSTRIAL category (but for no instances from the other categories). For instances on which the feature computation timed out, the backup solver was used.

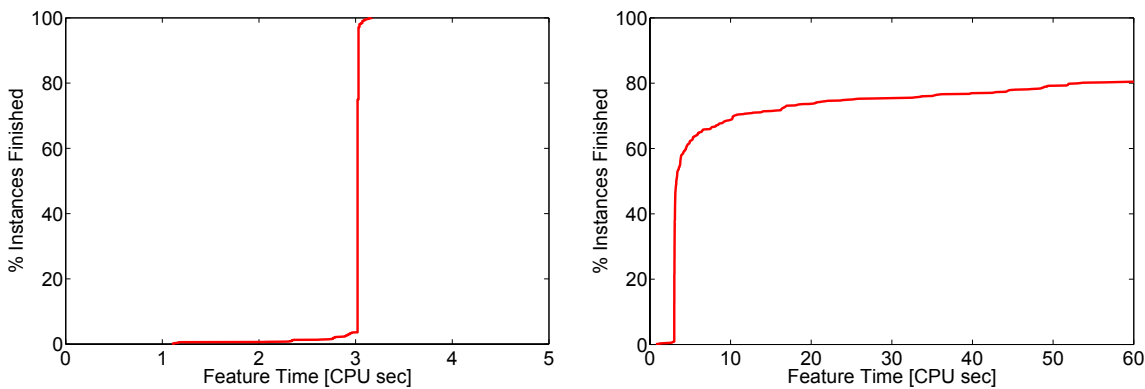

Figure 3: *Variability in feature computation times. The y-axis denotes the percentage of instances for which feature computation finished in at most the time given by the x-axis. Left:* RANDOM, *right:* INDUSTRIAL. *Note the different scales of the x-axes.*

Figure 3 illustrates this variation in feature computation time. For category RANDOM (left pane), feature computation never took significantly longer than three CPU seconds. In contrast, for category INDUSTRIAL, there was a fairly high variation, with the feature computation reaching the cut-off time of 60 CPU seconds for 19% of the instances. The average total feature computation times for categories RANDOM, HANDMADE, and INDUSTRIAL were 3.01, 4.22, and 14.4 CPU seconds, respectively.

### 4.3 RANDOM Category

For each category, we evaluated our SATzilla portfolios by comparing them against the best solvers from Table 1 for the respective category. Note that the solvers in this table are exactly the candidate solvers used in SATzilla.

Figure 4 shows the performance of SATzilla07 and the top three single solvers from Table 1 on category RANDOM. Note that we count the runtime of pre-solving, as well as the feature computation time, as part of SATzilla's runtime. Oracle(S) provides an upper bound on the performance that could be achieved by SATzilla: its runtime is that of a hypothetical version of SATzilla that makes every decision in an optimal fashion, and without any time spent computing features. Furthermore, it can also choose not to run the pre-solvers on an instance. Essentially, Oracle(S) thus indicates the performance that would be achieved by only running the best algorithm for each single instance. In Figure 4 (right), the horizontal line near the bottom of the plot shows the time SATzilla07(S,D$_r$) allots to pre-solving and (on average) to feature computation.

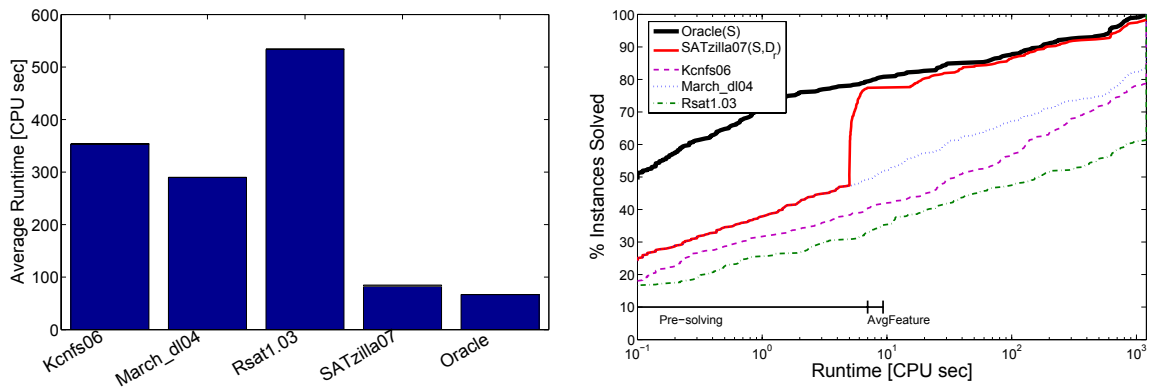

Figure 4: *Left: Average runtime, right: runtime cumulative distribution function (CDF) for different solvers on* RANDOM; *the average feature computation time was 2.3 CPU seconds (too insignificant to be visible in* SATzilla07's *runtime bar). All other solvers' CDFs are below the ones shown here (i.e., at each given runtime the maximum of the CDFs for the selected solvers is an upper bound for the CDF of any of the solvers considered in our experiments).*

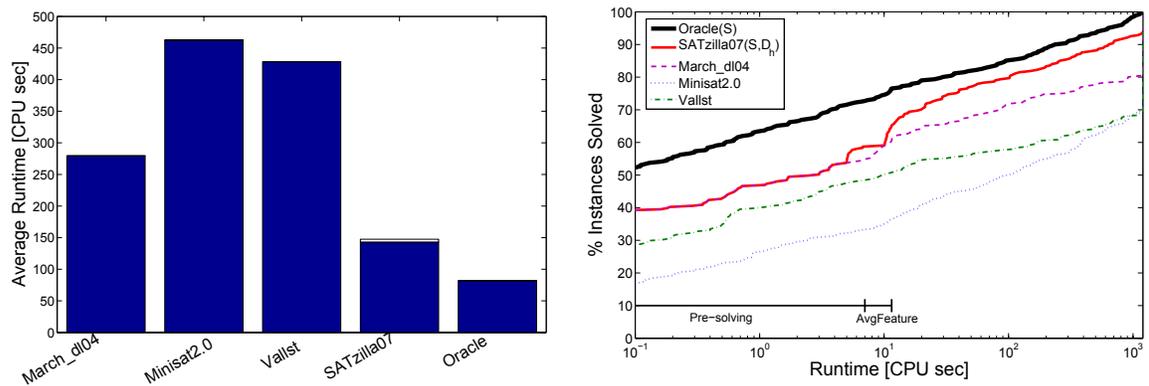

Figure 5: *Left: Average runtime, right: runtime CDF for different solvers on* HANDMADE; *the average feature computation time was 4.5 CPU seconds (shown as a white box on top of* SATzilla07's *runtime bar). All other solvers' CDFs are below the ones shown here.*

Overall, SATzilla07(S, D$_r$) achieved very good performance on data set RANDOM: It was more than three times faster on average than its best component solver, March_dl04 (see Figure 4, left), and also dominated it in terms of fraction of instances solved, solving 20% more instances within the cutoff time (see Figure 4, left). The runtime CDF plot also shows that the local-search-based pre-solver SAPS helped considerably by solving more than 20% of instances within 2 CPU seconds (this is reflected in the sharp increase in solved instances just before feature computation begins).

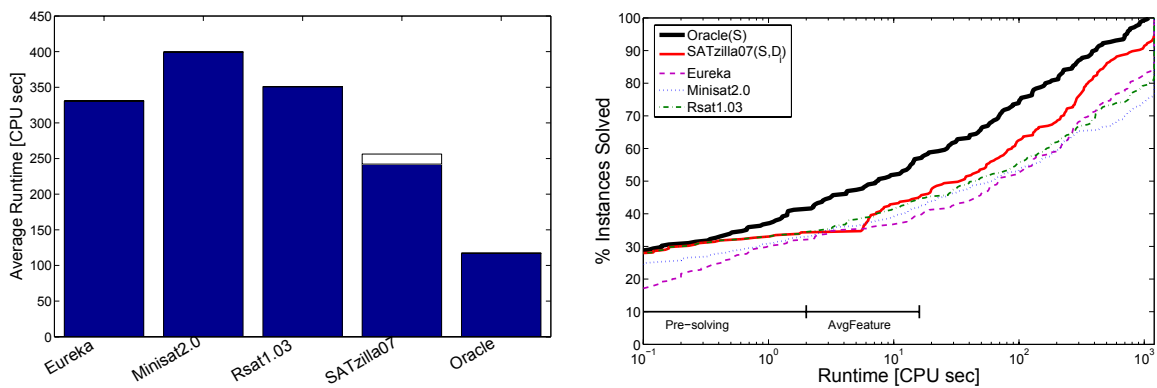

Figure 6: *Left: Average runtime, right: runtime CDF for different solvers on* INDUSTRIAL; *the average feature computation time was 14.1 CPU seconds (shown as a white box on top of* SATzilla07*'s runtime bar). All other solvers' CDFs are below the ones shown here.*

### 4.4 HANDMADE Category

SATzilla07's performance results for the HANDMADE category were also very good. Using the five component solvers listed in Table 3, its average runtime was about 45% less than that of the best single component solver (see Figure 5, left). The CDF plot in Figure 5 (right) shows that SATzilla07 dominated all its components and solved 13% more instances than the best non-portfolio solver.

### 4.5 INDUSTRIAL Category

We performed the same experiment for INDUSTRIAL instances as for the other categories in order to study SATzilla07's performance compared to its component solvers. SATzilla07 was more than 23% faster on average than the best component solver, Eureka (see Figure 6 (left)). Moreover, Figure 6 (right) shows that SATzilla07 also solved 9% more instances than Eureka within the cutoff time of 1 200 CPU seconds. Note that in this category, the feature computation timed out on 15.5% of the test instances after 60 CPU seconds; Eureka was used as a backup solver in those cases.

### 4.6 ALL

For our final category, ALL, a heterogeneous category that included the instances from all the above categories, a portfolio approach is especially appealing. SATzilla07 performed very well in this category, with an average runtime of less than half that of the best single solver, March_dl04 (159 vs. 389 CPU seconds, respectively). It also solved 20% more instances than any non-portfolio solver within the given time limit of 1 200 CPU seconds (see Figure 7).

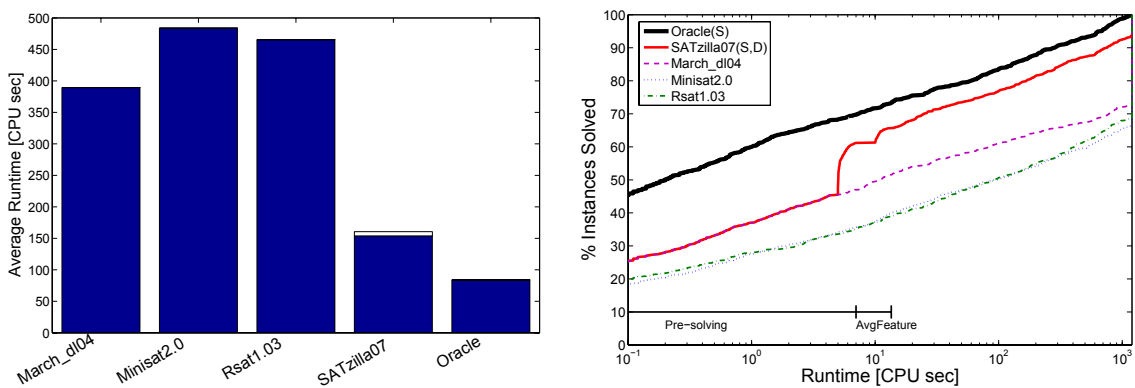

Figure 7: *Left: Average runtime; right: runtime CDF for different solvers on* `ALL`*; the average feature computation time was 6.7 CPU seconds (shown as a white box on top of* `SATzilla07`*'s runtime bar). All other solvers' CDFs are below the ones shown here.*

## 4.7 Classifier Accuracy

The satisfiability status classifiers trained on the various data sets were surprisingly effective in predicting satisfiability of `RANDOM` and `INDUSTRIAL` instances, where they reached classification accuracies of 94% and 92%, respectively. For `HANDMADE` and `ALL`, the classification accuracy was still considerably better than random guessing, at 70% and 78%, respectively.

Interestingly, our classifiers more often misclassified unsatisfiable instances as SAT than satisfiable instances as UNSAT. This effect can be seen from the confusion matrices in Table 6; it was most pronounced in the `HANDMADE` category, where the overall classification quality was also lowest: 40% of the `HANDMADE` instances that were classified as SAT are in fact unsatisfiable, while only 18% of the instances that were classified as UNSAT are in fact satisfiable.

## 5. Design II: `SATzilla` Beyond 2007

Despite `SATzilla07`'s success in the 2007 SAT Competition, there was still room for improvement. This section describes a number of design enhancements over `SATzilla07` that underly the new `SATzilla` versions, `SATzilla07`[+] and `SATzilla07`[*], which we describe in detail in Section 6 and evaluate experimentally in Section 7.

### 5.1 Automatically Selecting Pre-solvers

In `SATzilla07`, we identified pre-solvers and their cutoff times manually. There are several limitations to this approach. First and foremost, manual pre-solver selection does not scale well. If there are many candidate solvers, manually finding the best combination of pre-solvers and cutoff times can be difficult and requires significant amounts of valuable human time. In addition, the manual pre-solver selection we performed for `SATzilla07` concentrated solely on solving a large number of instances quickly and did not take into

|  | satisfiable | unsatisfiable |
|---|---|---|
| classified SAT | 91% | 9% |
| classified UNSAT | 5% | 95% |

RANDOM data set

|  | satisfiable | unsatisfiable |
|---|---|---|
| classified SAT | 60% | 40% |
| classified UNSAT | 18% | 82% |

HANDMADE data set

|  | satisfiable | unsatisfiable |
|---|---|---|
| classified SAT | 81% | 19% |
| classified UNSAT | 5% | 95% |

INDUSTRIAL data set

|  | satisfiable | unsatisfiable |
|---|---|---|
| classified SAT | 65% | 35% |
| classified UNSAT | 12% | 88% |

ALL data set

Table 6: *Confusion matrices for the satisfiability status classifier on data sets RANDOM, HANDMADE, INDUSTRIAL and ALL.*

account the pre-solvers' effect on model learning. In fact, there are three consequences of pre-solving.

1. Pre-solving solves some instances quickly before features are computed. In the context of the SAT competition, this improves SATzilla's scores for easy problem instances due to the "speed purse" component of the SAT competition scoring function. (See Section 5.3 below.)

2. Pre-solving increases SATzilla's runtime on instances not solved during pre-solving by adding the pre-solvers' time to every such instance. Like feature computation itself, this additional cost reduces SATzilla's scores.

3. Pre-solving filters out easy instances, allowing our empirical hardness models to be trained exclusively on harder instances.

While we considered (1) and (2) in our manual selection of pre-solvers, we did not consider (3), namely the fact that the use of different pre-solvers and/or cutoff times results in different training data and hence in different learned models, which can also affect a portfolio's effectiveness.

Our new automatic pre-solver selection technique works as follows. We committed in advance to using a maximum of two pre-solvers: one of three complete search algorithms and one of three local search algorithms. The three candidates for each of the search approaches are automatically determined for each data set as those with highest score on the validation set when run for a maximum of 10 CPU seconds. We also use a number of possible cutoff times, namely 2, 5 and 10 CPU seconds, as well as 0 seconds (i.e., the pre-solver is not run at all) and consider both orders in which the two pre-solvers can be run. For each of the resulting 288 possible combinations of two pre-solvers and cutoff times, SATzilla's performance on the validation data is evaluated by performing steps 6, 7 and 8 of the general methodology presented in Section 2:

6. determine the backup solver for selection when features time out;

7. construct an empirical hardness model for each algorithm; and

8. automatically select the best subset of algorithms to use as part of `SATzilla`.

The best-performing subset found in this last step—evaluated on validation data—is selected as the algorithm portfolio for the given combination of pre-solver / cutoff time pairs. Overall, this method aims to choose the pre-solver configuration that yields the best-performing portfolio.

## 5.2 Randomized Solver Subset Selection for a Large Set of Component Solvers

Our methodology from Section 2 relied on exhaustive subset search for choosing the best combination of component solvers. For a large number of component solvers, this is impossible ($N$ component solvers would require the consideration of $2^N$ solver sets, for each of which a model would have to be trained). The automatic pre-solver selection methods described in Section 5.1 above further worsens this situation: solver selection must be performed for every candidate configuration of pre-solvers, because new pre-solver configurations induce new models.

As an alternative to exhaustively considering all subsets, we implemented a randomized iterative improvement procedure to search for a good subset of solvers. The local search neighbourhood used by this procedure consists of all subsets of solvers that can be reached by adding or dropping a single component solver. Starting with a randomly selected subset of solvers, in each search step, we consider a neighbouring solver subset selected uniformly at random and accept it if validation set performance increases; otherwise, we accept the solver subset anyway with a probability of 5%. Once 100 steps have been performed with no improving step, a new run is started by re-initializing the search at random. After 10 such runs, the search is terminated and the best subset of solvers encountered during the search process is returned. Preliminary evidence suggests that this local search procedure efficiently finds very good subsets of solvers.

## 5.3 Predicting Performance Score Instead of Runtime

Our general portfolio methodology is based on empirical hardness models, which predict an algorithm's runtime. However, one might not simply be interested in using a portfolio to pick the solver with the lowest expected *runtime*. For example, in the SAT competition, solvers are evaluated based on a complex scoring function that depends only partly on a solver's runtime. Although the idiosyncrasies of this scoring function are somewhat particular to the SAT competition, the idea that a portfolio should be built to optimize a performance score more complex than runtime has wide applicability. In this section we describe techniques for building models that predict such a *performance score* directly.

One key issue is that—as long as we depend on standard supervised learning methods that require independent and identically distributed training data—we can only deal easily with scoring functions that actually associate a score with each single instance and combine the partial scores of all instances to compute the overall score. Given training data labeled with such a scoring function, `SATzilla` can simply learn a model of the score (rather than runtime) and then choose the solver with highest predicted score. Unfortunately, the scoring function used in the 2007 SAT Competition does not satisfy this independence property: the score a solver attains for solving a given instance depends in part on its (and, indeed, other solvers') performance on other, similar instances. More specifically, in the SAT competition each instance $P$ has a solution purse $SolutionP$ and a speed purse $SpeedP$; all instances in a given series (typically 5–40 similar instances) share one series purse $SeriesP$. Algorithms are ranked by summing three partial scores derived from these purses.

1. For each problem instance $P$, its solution purse is equally distributed between the solvers $S_i$ that solve the instance within the cutoff time (thereby rewarding robustness of a solver).

2. The speed purse for $P$ is divided among a set of solvers $S$ that solved the instance as $Score(P, S_i) = \frac{SpeedP \times SF(P,S_i)}{\Sigma_j SF(P,S_j)}$, where the speed factor $SF(P,S) = \frac{timeLimit(P)}{1+timeUsed(P,S)}$ is a measure of speed that discounts small absolute differences in runtime.

3. The series purse for each series is divided equally and distributed between the solvers $S_i$ that solved at least one instance in that series.[5]

$S_i$'s partial score from problem $P$'s solution and speed purses solely depends on the solver's own runtime for $P$ and the runtime of all competing solvers for $P$. Thus, given the runtimes of all competing solvers as part of the training data, we can compute the score contributions from the solution and speed purses of each instance $P$, and these two components are independent across instances. In contrast, since a solver's share of the series purse will depend on its performance on other instances in the series, its partial score received from the series purse for solving one instance is not independent of its performance on other instances.

Our solution to this problem is to approximate an instance's share of the series purse score by an independent score. If $N$ instances in a series are solved by any of SATzilla's component solvers, and if $n$ solvers solve at least one of the instances in that series, we assign a partial score of $SeriesP/(N \times n)$ to each solver $S_i$ (where $i = 1, \ldots, n$) for each instance in the series it solved. This approximation of a non-independent score as independent is not always perfect, but it is conservative because it defines a lower-bound on the partial score from the series purse. Predicted scores will only be used in SATzilla to choose between different solvers on a per-instance basis. Thus, the partial score of a solver for an instance should reflect how much it would contribute to SATzilla's score. If SATzilla were perfect (i.e., for each instance, it always selected the best algorithm) our score approximation would be correct: SATzilla would solve all $N$ instances from the series that any component solver can solve, and thus would actually achieve the series score $SeriesP/(N \times n) \times N = SeriesP/n$. If SATzilla performed very poorly and did not solve any instance in the series, our approximation would also be exact, since it would estimate the partial series score as zero. Finally, if SATzilla were to pick successful solvers for some (say, $M$) but not all instances of the series that can be solved by its component solvers (i.e., $M < N$), we would underestimate the partial series purse, since $SeriesP/(N \times n) \times M < SeriesP/n$.

---

5. See http://www.satcompetition.org/2007/rules07.html for further details.

While our learning techniques require an approximation of the performance score as an independent score, our experimental evaluations of solvers' scores employ the actual SAT competition scoring function. As explained previously, in the SAT competition, the performance score of a solver depends on the score of all other solvers in the competition. In order to simulate a competition, we select a large number of solvers and pretend that these "reference solvers" and `SATzilla` are the only solvers in the competition; throughout our analysis we use the 19 solvers listed in Tables 1, 9 and 10. This is not a perfect simulation, since the scores change somewhat when different solvers are added to or removed from the competition. However, we obtain much better approximations of performance score by following the methodology outlined here than by using cruder measures, such as learning models to predict mean runtime or the numbers of benchmark instances solved.

Finally, predicting performance score instead of runtime has a number of implications for the components of `SATzilla`. First, notice that we can compute an exact score for each algorithm and instance, even if the algorithm times out unsuccessfully or crashes—in these cases, the score from all three components is simply zero. When predicting scores instead of runtimes, we thus do not need to rely on censored sampling techniques (see Section 2.2) anymore. Secondly, notice that the oracles for maximizing SAT competition score and for minimizing runtime are identical, since always using the solver with the smallest runtime guarantees that the highest values in all three components are obtained.

## 5.4 Introducing Local Search Solvers into `SATzilla`

SAT solvers based on local search are well known to be effective on certain classes of satisfiable instances. In fact, there are classes of hard random satisfiable instances that only local search solvers can solve in a reasonable amount of time (Hoos & Stützle, 2005). However, all high-performance local-search-based SAT solvers are incomplete and cannot solve unsatisfiable instances. In previous versions of `SATzilla` we avoided using local search algorithms because of the risk that we would select them for unsatisfiable instances, where they would run uselessly until reaching the cutoff time.

When we shift to predicting and optimizing performance score instead of runtime, this issue turns out not to matter anymore. Treating every solver as a black box, local search solvers always get a score of exactly zero on unsatisfiable instances since they are guaranteed not to solve them within the cutoff time. (Of course, they do not need to be run on an instance during training if the instance is known to be unsatisfiable.) Hence, we can build models for predicting the score of local search solvers using exactly the same methods as for complete solvers.

## 5.5 More General Hierarchical Hardness Models

Our benchmark set `ALL` consists of all instances from the categories `RANDOM`, `HANDMADE` and `INDUSTRIAL`. In order to further improve performance on this very heterogeneous instance distribution, we extend our previous hierarchical hardness model approach (predicting satisfiability status and then using a mixture of two conditional models) to the more general scenario of six underlying empirical hardness models (one for each combination of category and satisfiability status). The output of the general hierarchical model is a linear weighted combination of the output of each component. As described in Section 2.3, we can approx-imate the model selection oracle by a softmax function whose parameters are estimated using EM.

|  | "Old" instances from before 2007 | "New" instances from 2007 |
|---|---|---|
| Training (40%) | $T_o$ (1 925 instances) | $T_n$ (347 instances) |
| Validation (30%) | $V_o$ (1 443 instances) | $V_n$ (261 instances) |
| Test (30%) | $E_o$ (1 443 instances) | $E_n$ (261 instances) |

Table 7: *Instances from before 2007 and from 2007 randomly split into training (T), validation (V) and test (E) data sets. These sets include instances for all categories:* RANDOM, HANDMADE *and* INDUSTRIAL.

| Data set | Training | Validation | Test |
|---|---|---|---|
| $D$ | $T_o$ | $V_o$ | $E_o$ |
| $D'$ | $T_o$ | $V_o$ | $E_o \cup E_n$ |
| $D^+$ | $T_o \cup T_n$ | $V_o \cup V_n$ | $E_o \cup E_n$ |

Table 8: *Data sets used in our experiments. $D$ was used in our first series of experiments in Section 4, $D'$ and $D^+$ are used in our second series of experiments. Note that data sets $D$ and $D'$ use identical training and validation data, but different test data.*

## 6. Construction II: Building the Improved SATzilla Versions

In this section we describe the construction of new SATzilla versions that incorporate new design elements from the previous section. We also describe two versions based on our old design, which we use to evaluate the impact of our changes.

### 6.1 Benchmark Instances

In addition to all instances used in Section 3.1, we added the 869 instances from the 2007 SAT Competition into our four data sets. Overall, this resulted in 5 680 instances: 2 811 instances in category RANDOM, 1 676 in category HANDMADE and 1 193 in category INDUSTRIAL. Recall that in Section 3.1 we dropped instances that could not be solved by any of the seven solvers in Table 1. We follow the same methodology here, but extend our solver set by the 12 solvers in Tables 9 and 10. Now, 71.8% of the instances can be solved by at least one of our 19 solvers within the cutoff time of 1 200 CPU seconds on our reference machine; the remaining instances were excluded from our analysis.

We randomly split the above benchmark sets into training, validation and test sets, as described in Table 7. All parameter tuning and intermediate testing was performed on validation sets, and test sets were used only to generate the final results reported here.

We will be interested in analyzing SATzilla's performance as we vary the data that was used to train it. To make it easy to refer to our different data sets, we describe them here and assign names ($D$, $D'$, $D^+$) to them. Table 7 shows the division of our data into "old" (pre-2007) and "new" (2007) instances. Table 8 shows how we combined this data to construct the three data sets we use for evaluation. Data set $D$ is the one introduced and used in Section 3.1: it uses only pre-2007 instances for training, validation and testing. Data set $D'$ uses the same training and validation data sets, but differs in its test sets, which include both old and new instances. Data set $D^+$ combines both old and new instances in its training, validation *and* test sets.

| Solver | Reference |
|---|---|
| `Kcnfs04` | Dequen and Dubois (2007) |
| `TTS` | Spence (2007) |
| `Picosat` | Biere (2007) |
| `MXC` | Bregman and Mitchell (2007) |
| `March_ks` | Heule and v. Maaren (2007) |
| `TinisatElite` | Huang (2007) |
| `Minisat07` | Sörensson and Eén (2007) |
| `Rsat 2.0` | Pipatsrisawat and Darwiche (2007) |

Table 9: Eight complete solvers from the 2007 SAT Competition.

| Solver | Reference |
|---|---|
| `Ranov` | Pham and Anbulagan (2007) |
| `Ag2wsat0` | C. M. Li and Zhang (2007) |
| `Ag2wsat+` | Wei, Li, and Zhang (2007) |
| `Gnovelty+` | Pham and Gretton (2007) |

Table 10: Four local search solvers from the 2007 SAT Competition.

Thus, note that data sets $D'$ and $D^+$ use the same test sets, meaning that the performance of portfolios trained using these different sets can be compared directly. However, we expect a portfolio trained using $D^+$ to be at least slightly better, because it has access to more data. As before, when we want to refer to only the `RANDOM` instances from $D^+$, we write $D_r^+$; likewise, we write $D_h^+$ for `HANDMADE`, $D_i^+$ for `INDUSTRIAL`, etc.

## 6.2 Extending the Set of Component Solvers

In addition to the seven "old" solvers used in `SATzilla07` (previously described in Table 1), we considered eight new complete solvers and four local search solvers from the 2007 SAT Competition for inclusion in our portfolio; these solvers are described in Tables 9 and 10.

As with training instances, we treat sets of candidate solvers as an input parameter of `SATzilla`. The sets of candidate solvers used in our experiments are detailed in Table 11.

## 6.3 Different `SATzilla` Versions

Having just introduced new design ideas for `SATzilla` (Section 5), new training data (Section 6.1) and new solvers (Section 6.2), we were interested in evaluating how much our portfolio improved as a result. In order to gain insights into how much performance improvement was achieved by these different changes, we studied several intermediate SATzilla solvers, which are summarized in Table 12.

| Name of Set | Solvers in the Set |
|---|---|
| S | all 7 solvers from Table 1 |
| S$^+$ | all 15 solvers from Tables 1 and 9 |
| S$^{++}$ | all 19 solvers from Tables 1, 9 and 10 |

Table 11: Solver sets used in our second series of experiments.

| SATzilla version | Description |
|---|---|
| SATzilla07(S,D′) | This is the version we entered into the 2007 SAT Competition (Section 3), but evaluated on an extended test set. |
| SATzilla07(S$^+$,D$^+$) | This version is built using the same design as described in Section 3, but includes new complete solvers (Table 9) and new data (Section 6.1). |
| SATzilla07$^+$(S$^{++}$,D$^+$) | In addition to new complete solvers and data, this version uses local search solvers (Table 10) and all of the new design elements from Section 5 except "more general hierarchical hardness models" (Section 5.5). |
| SATzilla07$^*$(S$^{++}$,D$^+$) | This version uses all solvers, all data and all new design elements. Unlike for the other versions, we trained only one variant of this solver for use in all data set categories. |

Table 12: The different SATzilla versions evaluated in our second set of experiments.

Observe that all of these solvers were built using identical test data and were thus directly comparable. We generally expected each solver to outperform its predecessors in the list. The exception was SATzilla07$^*$(S$^{++}$,D$^+$): instead of aiming for increased performance, this last solver was designed to achieve good performance across a broader range of instances. Thus, we expected SATzilla07$^*$(S$^{++}$,D$^+$) to outperform the others on category ALL, but not to outperform SATzilla07$^+$(S$^{++}$,D$^+$) on the more specific categories.

### 6.4 Constructing SATzilla07$^+$(S$^{++}$,D$^+$) and SATzilla07$^*$(S$^{++}$,D$^+$)

The construction of SATzilla07(S,D) was already described in Section 3; SATzilla07(S,D') differed in the test set we used to evaluate it, but was otherwise identical. The construction of SATzilla07(S$^+$,D$^+$) was the same as that for SATzilla07(S,D'), except that it relied on different solvers and corresponding training data. SATzilla07$^+$(S$^{++}$,D$^+$) and SATzilla07$^*$(S$^{++}$,D$^+$) incorporated the new techniques introduced in Section 5. In this section we briefly describe how these solvers were constructed.

We used the same set of features as for SATzilla07 (see Section 3.3). We also used the same execution environment and cutoff times. Pre-solvers were identified automatically as described in Section 5.1, using the (automatically determined) candidate solvers listed in Table 13. The final sets of pre-solvers selected for each version of SATzilla are listed in Section 7 (Tables 14, 17, 20 and 23). Based on the solvers' scores on validation data sets, we automatically determined the backup solvers for RANDOM, HANDMADE, INDUSTRIAL and ALL to be March_ks, March_dl04, Eureka and Eureka, respectively.

|  | RANDOM | HANDMADE | INDUSTRIAL | ALL |
|---|---|---|---|---|
| **Complete** **Pre-solver** **Candidates** | Kcnfs06 March_dl04 March_ks | March_dl04 Vallst March_ks | Rsat 1.03 Picosat Rsat 2.0 | Minisat07 March_ks March_dl04 |
| **Local Search** **Pre-solver** **Candidates** | Ag2wsat0 Gnovelty+ SAPS | Ag2wsat0 Ag2wsat+ Gnovelty+ | Ag2wsat0 Ag2wsat+ Gnovelty+ | SAPS Ag2wsat0 Gnovelty+ |

Table 13: Pre-solver candidates for our four data sets. These candidates were automatically chosen based on the scores on validation data achieved by running the respective algorithms for a maximum of 10 CPU seconds.

We built models to predict the performance score of each algorithm. This score is well defined even in case of timeouts and crashes; thus, there was no need to deal with censored data. Like SATzilla07, SATzilla07$^+$ used hierarchical empirical hardness models (Xu et al., 2007a) with two underlying models ($M_{sat}$ and $M_{unsat}$) for predicting a solver's score. For SATzilla07*, we built more general hierarchical hardness models for predicting scores as described in Section 5.5; these models were based on six underlying empirical hardness models ($M_{sat}$ and $M_{unsat}$ trained on data from each SAT competition category).

We chose solver subsets based on the results of our local search procedure for subset search as outlined in Section 5.2. The resulting final components of SATzilla07, SATzilla07$^+$ and SATzilla07* for each category are described in detail in the following section.

## 7. Evaluation II: Performance Analysis of the Improved SATzilla Versions

In this section, we investigate the effectiveness of our new techniques by evaluating the four SATzilla versions listed in Table 12: SATzilla07(S,D'), SATzilla07(S$^+$,D$^+$), SATzilla07$^+$(S$^{++}$,D$^+$) and SATzilla07*(S$^{++}$,D$^+$). To evaluate their performance, we constructed a simulated SAT competition using the same scoring function as in the 2007 SAT Competition, but differing in a number of important aspects. The participants in our competition were the 19 solvers listed in Tables 1, 9, and 10 (all solvers were considered for all categories), and the test instances were $E_o \cup E_n$ as described in Tables 7 and 8. Furthermore, our computational infrastructure (see Section 3.4) differed from the 2007 competition, and we also used shorter cutoff times of 1200 seconds. For these reasons some solvers ranked slightly differently in our simulated competition than in the 2007 competition.

### 7.1 RANDOM Category

Table 14 shows the configuration of the three different SATzilla versions designed for the RANDOM category. Note that the automatic solver selection in SATzilla07$^+$(S$^{++}$,D$_r^+$) included different solvers than the ones used in SATzilla07(S$^+$,D$_r^+$); in particular, it chose three local search solvers, `Ranov`, `Ag2wsat+`, and `Gnovelty+`, that were not available to `SATzilla07`. Also, the automatic pre-solver selection chose a different order and cutoff time of pre-solvers than our manual selection: it chose to first run `SAPS` for two CPU seconds, followed by two CPU seconds of `Kcnfs06`. Even though running the local search algorithm `SAPS` did not help for solving unsatisfiable instances, we see in Figure 8 (left) that `SAPS` solved many more instances than `March_dl04` in the first few seconds.

| SATzilla version | Pre-Solvers (time) | Component solvers |
|---|---|---|
| SATzilla07(S,D'$_r$) | March_dl04(5); SAPS(2) | Kcnfs06, March_dl04, Rsat 1.03 |
| SATzilla07(S$^+$,D$_r^+$) | March_dl04(5); SAPS(2) | Kcnfs06, March_dl04, March_ks, Minisat07 |
| SATzilla07$^+$(S$^{++}$,D$_r^+$) | SAPS(2); Kcnfs06(2) | Kcnfs06, March_ks, Minisat07, Ranov, Ag2wsat+, Gnovelty+ |

Table 14: *SATzilla's configuration for the RANDOM category; cutoff times for pre-solvers are specified in CPU seconds.*

Table 15 shows the performance of different versions of `SATzilla` compared to the best solvers in the `RANDOM` category. All versions of `SATzilla` outperformed every non-portfolio solver in terms of average runtime and number of instances solved. `SATzilla07$^+$` and `SATzilla07*`, the variants optimizing score rather than another objective function, also clearly achieved higher scores than the non-portfolio solvers. This was not always the case for the other versions; for example, `SATzilla07(S$^+$,D$_r^+$)` achieved only 86.6% of the score of the best solver, `Gnovelty+` (where scores were computed based on a reference set of 20 reference solvers: the 19 solvers from Tables 1, 9, and 10, as well as `SATzilla07(S$^+$,D$_r^+$)`). Table 15 and Figure 8 show that adding complete solvers and training data did not improve `SATzilla07` much. At the same time, substantial improvements were achieved by the new mechanisms in `SATzilla07$^+$`, leading to 11% more instances solved, a reduction of average runtime by more than half, and an increase in score by over 50%. Interestingly, the performance of the more general `SATzilla07*(S$^{++}$,D$^+$)` trained on instance mix `ALL` and tested on the `RANDOM` category was quite close to the best version of `SATzilla` specifically designed for `RANDOM` instances, `SATzilla07$^+$(S$^{++}$,D$_r^+$)`. Note that due to their excellent performance on satisfiable instances, the local search solvers in Table 15 (`Gnovelty+` and `Ag2wsat` variants) tended to have higher overall scores than the complete solvers (`Kcnfs04` and `March_ks`) even though they solved fewer instances and in particular could not solve any unsatisfiable instance. In the 2007 SAT Competition, however, all winners of the random SAT+UNSAT category were complete solvers, which lead us to speculate that local search solvers were not considered in this category (in the random SAT category, all winners were indeed local search solvers).

Figure 8 presents CDFs summarizing the performance of the best non-portfolio solvers, `SATzilla` solvers and two oracles. All non-portfolio solvers omitted had CDFs below those shown. As in Section 4, the oracles represent ideal versions of `SATzilla` that choose among component solvers perfectly and without any computational cost. More specifically, given an instance, an oracle picks the fastest algorithm; it is allowed to consider `SAPS` (with a maximum runtime of 10 CPU seconds) and any solver from the given set (S for one oracle and $S^{++}$ for the other).

| Solver | Avg. runtime [s] | Solved [%] | Performance score |
|---|---|---|---|
| Kcnfs04 | 852 | 32.1 | 38309 |
| March_ks | **351** | **78.4** | 113666 |
| Ag2wsat0 | 479 | 62.0 | 119919 |
| Ag2wsat+ | 510 | 59.1 | 110218 |
| Gnovelty+ | 410 | 67.4 | **131703** |
| SATzilla07(S,$D'_r$) | 231 | 85.4 | — (86.6%) |
| SATzilla07($S^+$,$D^+_r$) | 218 | 86.5 | — (88.7%) |
| SATzilla07$^+$($S^{++}$,$D^+_r$) | **84** | **97.8** | **189436 (143.8%)** |
| SATzilla07$^*$($S^{++}$,$D^+$) | 113 | 95.8 | — (137.8%) |

Table 15: *The performance of* SATzilla *compared to the best solvers on* RANDOM. *The cutoff time was 1 200 CPU seconds;* SATzilla07$^*$($S^{++}$,$D^+$) *was trained on* ALL. *Scores were computed based on 20 reference solvers: the 19 solvers from Tables 1, 9, and 10, as well as one version of* SATzilla. *To compute the score for each non-*SATzilla *solver, the* SATzilla *version used as a member of the set of reference solvers was* SATzilla07$^+$($S^{++}$,$D^+_r$). *Since we did not include* SATzilla *versions other than* SATzilla07$^+$($S^{++}$,$D^+_r$) *in the set of reference solvers, scores for these solvers are incomparable to the other scores given here, and therefore we do not report them. Instead, for each* SATzilla *solver, we indicate in parentheses its performance score as a percentage of the highest score achieved by a non-portfolio solver, given a reference set in which the appropriate* SATzilla *solver took the place of* SATzilla07$^+$($S^{++}$,$D^+_r$).*

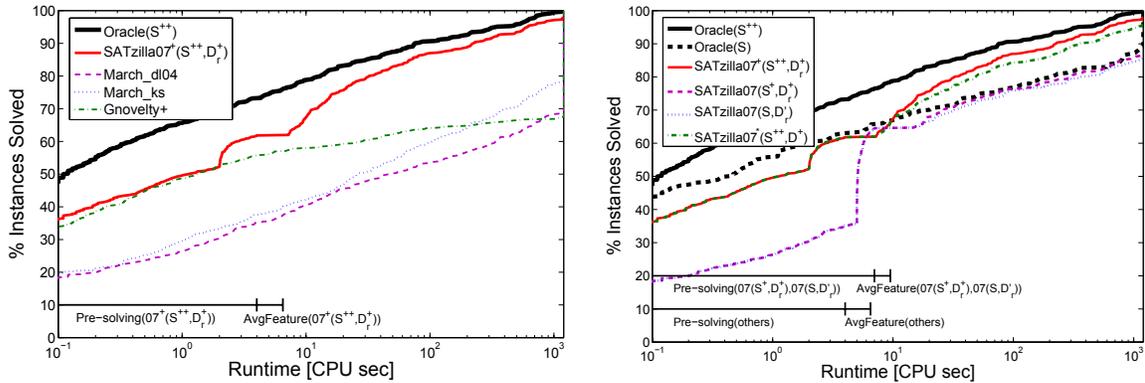

Figure 8: *Left: CDFs for* SATzilla07$^+$($S^{++}$,$D^+_r$) *and the best non-portfolio solvers on* RANDOM; *right: CDFs for the different versions of* SATzilla *on* RANDOM *shown in Table 14, where* SATzilla07$^*$($S^{++}$,$D^+$) *was trained on* ALL. *All other solvers' CDFs are below the ones shown here.*

Table 16 indicates how often each component solver of SATzilla07$^+$($S^{++}$,$D^+_r$) was selected, how often it was successful, and the amount of its average runtime. We found that the solvers picked by SATzilla07$^+$(S$^{++}$,D$_r^+$) solved the given instance in most cases. Another interesting observation is that when a solver's success ratio was high, its average runtime tended to be lower.

| Pre-Solver (Pre-Time) | Solved [%] | | Avg. Runtime [CPU sec] |
|---|---|---|---|
| SAPS(2) | 52.2 | | 1.1 |
| March_dl04(2) | 9.6 | | 1.68 |
| **Selected Solver** | **Selected [%]** | **Success [%]** | **Avg. Runtime [CPU sec]** |
| March_dl04 | 34.8 | 96.2 | 294.8 |
| Gnovelty+ | 28.8 | 93.9 | 143.6 |
| March_ks | 23.9 | 92.6 | 213.3 |
| Minisat07 | 4.4 | 100 | 61.0 |
| Ranov | 4.0 | 100 | 6.9 |
| Ag2wsat+ | 4.0 | 77.8 | 357.9 |

Table 16: *The solvers selected by SATzilla07$^+$(S$^{++}$,D$_r^+$) for the RANDOM category. Note that column "Selected [%]" shows the percentage of instances remaining after pre-solving for which the algorithm was selected, and this sums to 100%. Cutoff times for pre-solvers are specified in CPU seconds.*

## 7.2 HANDMADE Category

The configurations of the three SATzilla versions designed for the HANDMADE category are shown in Table 17. Again, SATzilla07$^+$(S$^{++}$,D$_h^+$) included three local search solvers, Ranov, Ag2wsat+ and Gnovelty+, which had not been available to SATzilla07. Like our manual choice in SATzilla07, the automatic pre-solver selection chose to run March_dl04 for five CPU seconds. Unlike the manual selection, it abstained from using SAPS (or indeed any other solver) as a second pre-solver. Table 18 shows the performance of the different versions of SATzilla compared to the best solvers for category HANDMADE. Here, about half of the observed performance improvement was achieved by using more solvers and more training data; the other half was due to the improvements in SATzilla07$^+$. Note that for the HANDMADE category, SATzilla07$^*$(S$^{++}$,D$^+$) performed quite poorly. We attribute this to a weakness of the feature-based classifier on HANDMADE instances, an issue we discuss further in Section 7.4.

Table 19 indicates how often each component solver of SATzilla07$^+$(S$^{++}$,D$_h^+$) was selected, how many problem instances it solved, and its average runtime for these runs. There are many solvers that SATzilla07$^+$(S$^{++}$,D$_h^+$) picked quite rarely; however, in most cases, their success ratios are close to 100%, and their average runtimes are very low.

## 7.3 INDUSTRIAL Category

Table 20 shows the configuration of the three different SATzilla versions designed for the INDUSTRIAL category. Local search solvers performed quite poorly for the instances in this category, with the best local search solver, Ag2wsat0, only solving 23% of the instances within the cutoff time. Consequently, no local search solver was selected by the automatic solver subset selection in `SATzilla07`[+]`(S`[++]`,D`$_i^+$`)`. However, automatic pre-solver selection did include the local search solver `Gnovelty+` as the second pre-solver, to be run for 2 CPU seconds after 10 CPU seconds of running `Rsat 2.0`.

| SATzilla | Pre-Solver (time) | Components |
|---|---|---|
| `SATzilla07(S,D'_h)` | `March_dl04(5)`; SAPS(2) | `Kcnfs06`, `March_dl04`, `Minisat 2.0`, `Rsat 1.03` |
| `SATzilla07(S^+,D_h^+)` | `March_dl04(5)`; SAPS(2) | `Vallst`, `Zchaff_rand`, `TTS`, `MXC`, `March_ks`, `Minisat07`, `Rsat 2.0` |
| `SATzilla07^+(S^{++},D_h^+)` | `March_ks(5)` | `Eureka`, `March_dl04`; `Minisat 2.0`, `Rsat 1.03`, `Vallst`, `TTS`, `Picosat`, `MXC`, `March_ks`, `TinisatElite`, `Minisat07`, `Rsat 2.0`, `Ranov`, `Ag2wsat0`, `Gnovelty+` |

Table 17: `SATzilla`'s configuration for the `HANDMADE` category.

| Solver | Avg. runtime [s] | Solved [%] | Performance score |
|---|---|---|---|
| TTS | 729 | 41.1 | 40669 |
| MXC | 527 | 61.9 | 43024 |
| `March_ks` | 494 | 63.9 | 68859 |
| `Minisat07` | 438 | 68.9 | 59863 |
| `March_dl04` | **408** | **72.4** | **73226** |
| `SATzilla07(S,D'_h)` | 284 | 80.4 | — (93.5%) |
| `SATzilla07(S^+,D_h^+)` | 203 | 87.4 | — (118.8%) |
| `SATzilla07^+(S^{++},D_h^+)` | **131** | **95.6** | **112287 (153.3%)** |
| `SATzilla07^*(S^{++},D^+)` | 215 | 88.0 | — (110.5%) |

Table 18: *The performance of `SATzilla` compared to the best solvers on `HANDMADE`. Scores for non-portfolio solvers were computed using a reference set in which the only `SATzilla` solver was `SATzilla07^+(S^{++},D_h^+)`. Cutoff time: 1200 CPU seconds; `SATzilla07^*(S^{++},D^+)` was trained on `ALL`.*

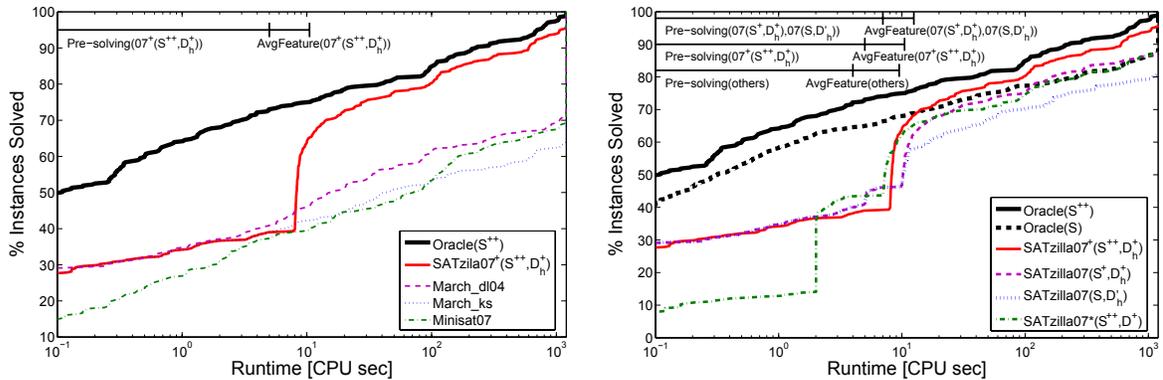

Figure 9: *Left: CDFs for `SATzilla07^+(S^{++},D_h^+)` and the best non-portfolio solvers on `HANDMADE`; right: CDFs for the different versions of `SATzilla` on `HANDMADE` shown in Table 17, where `SATzilla07^*(S^{++},D^+)` was trained on `ALL`. All other solvers' CDFs are below the ones shown here.*

| Pre-Solver (Pre-Time) | Solved [%] | | Avg. Runtime [CPU sec] |
|---|---|---|---|
| March_ks(5) | 39.0 | | 3.2 |
| **Selected Solver** | **Selected [%]** | **Success [%]** | **Avg. Runtime [CPU sec]** |
| Minisat07 | 40.4 | 89.3 | 205.1 |
| TTS | 11.5 | 91.7 | 133.2 |
| MXC | 7.2 | 93.3 | 310.5 |
| March_ks | 7.2 | 100 | 544.7 |
| Eureka | 5.8 | 100 | 0.34 |
| March_dl04 | 5.8 | 91.7 | 317.6 |
| Rsat 1.03 | 4.8 | 100 | 185.1 |
| Picosat | 3.9 | 100 | 1.7 |
| Ag2wsat0 | 3.4 | 100 | 0.5 |
| TinisatElite | 2.9 | 100 | 86.5 |
| Ranov | 2.9 | 83.3 | 206.1 |
| Minisat 2.0 | 1.4 | 66.7 | 796.5 |
| Rsat 2.0 | 1.4 | 100 | 0.9 |
| Gnovelty+ | 1.0 | 100 | 3.2 |
| Vallst | 0.5 | 100 | <0.01 |

Table 19: *The solvers selected by* `SATzilla07`[+] `(S`[++]`,D`$_h^+$`)` *for the* `HANDMADE` *category.*

| SATzilla | Pre-Solver (time) | Components |
|---|---|---|
| `SATzilla07(S,D'`$_i$`)` | Rsat 1.03 (2) | Eureka, March_dl04, Minisat 2.0, Rsat 1.03 |
| `SATzilla07(S`[+]`,D`$_i^+$`)` | Rsat 2.0 (2) | Eureka, March_dl04, Minisat 2.0, Zchaff_Rand, TTS, Picosat, March_ks |
| `SATzilla07`[+]`(S`[++]`,D`$_i^+$`)` | Rsat 2.0 (10); Gnovelty+(2) | Eureka, March_dl04, Minisat 2.0, Rsat 1.03, TTS, Picosat, Minisat07, Rsat 2.0 |

Table 20: `SATzilla`*'s configuration for the* `INDUSTRIAL` *category.*

Table 21 compares the performance of different versions of `SATzilla` and the best solvers on `INDUSTRIAL` instances. It is not surprising that more training data and more solvers helped `SATzilla07` to improve in terms of all our metrics (avg. runtime, percentage solved and score). A somewhat bigger improvement was due to the new mechanisms in `SATzilla07`[+] that led to `SATzilla07`[+]`(S`[++]`,D`$_i^+$`)` outperforming every non-portfolio solver with respect to every metric, specially in terms of performance score. Note that the general `SATzilla` version `SATzilla07`[*]`(S`[++]`,D`[+]`)` trained on `ALL` achieved performance very close to that of `SATzilla07`[+]`(S`[++]`,D`$_i^+$`)` on the `INDUSTRIAL` data set in terms of average runtime and percentage of solved instances.

| Solver | Avg. runtime [s] | Solved [%] | Performance score |
|---|---|---|---|
| Rsat 1.03 | 353 | 80.8 | 52740 |
| Rsat 2.0 | 365 | 80.8 | 51299 |
| Picosat | **282** | **85.9** | 66561 |
| TinisatElite | 452 | 70.8 | 40867 |
| Minisat07 | 372 | 76.6 | 60002 |
| Eureka | 349 | 83.2 | **71505** |
| SATzilla07(S,$D_i'$) | 298 | 87.6 | — (91.3%) |
| SATzilla07(S$^+$,$D_i^+$) | 262 | 89.0 | — (98.2%) |
| SATzilla07$^+$(S$^{++}$,$D_i^+$) | **233** | **93.1** | 79724 **(111.5%)** |
| SATzilla07$^*$(S$^{++}$,$D^+$) | 239 | 92.7 | — (104.8%) |

Table 21: *The performance of SATzilla compared to the best solvers on INDUSTRIAL. Scores for non-portfolio solvers were computed using a reference set in which the only SATzilla solver was SATzilla07$^+$(S$^{++}$,$D_i^+$). Cutoff time: 1200 CPU seconds; SATzilla07$^*$(S$^{++}$,$D^+$) was trained on ALL.*

| Pre-Solver (Pre-Time) | Solved [%] | | Avg. Runtime [CPU sec] |
|---|---|---|---|
| Rsat 2.0(10) | 38.1 | | 6.8 |
| Gnovelty+ (2) | 0.3 | | 2.0 |

| Selected Solver | Selected [%] | Success [%] | Avg. Runtime [CPU sec] |
|---|---|---|---|
| Eureka (BACKUP) | 29.1 | 88.5 | 385.4 |
| Eureka | 15.1 | 100 | 394.2 |
| Picosat | 14.5 | 96.2 | 179.6 |
| Minisat07 | 14.0 | 84.0 | 306.3 |
| Minisat 2.0 | 12.3 | 68.2 | 709.2 |
| March_dl04 | 8.4 | 86.7 | 180.8 |
| TTS | 3.9 | 100 | 0.7 |
| Rsat 2.0 | 1.7 | 100 | 281.6 |
| Rsat 1.03 | 1.1 | 100 | 10.6 |

Table 22: *The solvers selected by SATzilla07$^+$(S$^{++}$,$D_i^+$) for the INDUSTRIAL category.*

As can be seen from Figure 10, the performance improvements achieved by SATzilla over non-portfolio solvers are smaller for the INDUSTRIAL category than for other categories. Note that the best INDUSTRIAL solver performed very well, solving 85.9% of the instances within the cutoff time of 1200 CPU seconds.[6] Still, SATzilla07$^+$(S$^{++}$,$D_i^+$) had significantly smaller average runtime (17%) and solved 7.2% more instances than the best component solver, Picosat. Likewise, the score for SATzilla07$^+$(S$^{++}$,$D_i^+$) was 11.5% higher than that of the top-ranking component solver (in terms of score), Eureka.

---

6. Recall that this number means the solver solved 85.9% of the instances that could be solved by at least one solver. Compared to our other data sets, it seems that either solvers exhibited more similar behavior on INDUSTRIAL instances or that instances in this category exhibited greater variability in hardness.

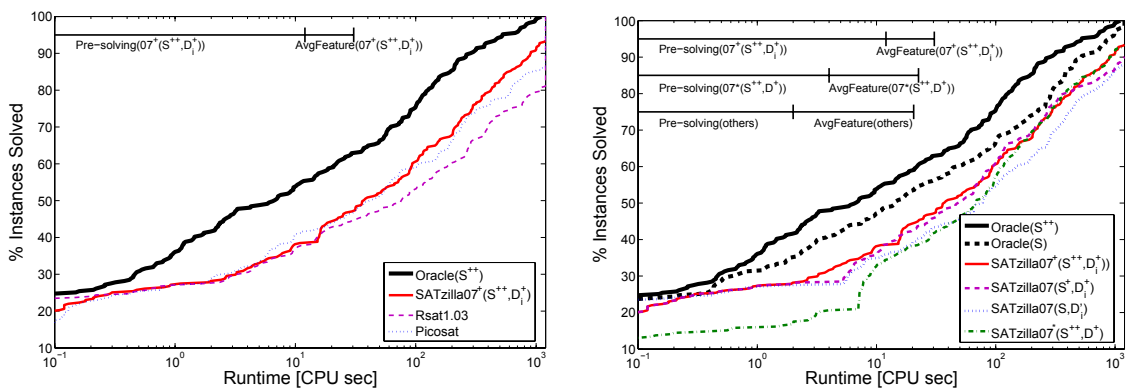

Figure 10: *Left: CDFs for `SATzilla07`$^+$`(S`$^{++}$`,D`$_i^+$`)` and the best non-portfolio solvers on `INDUSTRIAL`; right: CDFs for the different versions of `SATzilla` on `INDUSTRIAL` shown in Table 20, where `SATzilla07`$^*$` (S`$^{++}$`,D`$^+$`)` was trained on `ALL`. All other solvers' CDFs (including `Eureka`'s) are below the ones shown here.*

Table 22 indicates how often each component solver of `SATzilla07`$^+$`(S`$^{++}$`,D`$_i^+$`)` was selected, how many problem instances it solved, and its average runtime for these runs. In this case, backup solver `Eureka` was used for problem instances for which feature computation timed out and pre-solvers did not produce a solution.

### 7.4 `ALL`

There are four versions of `SATzilla` specialized for category `ALL`. Their detailed configurations are listed in Table 23. The results of automatic pre-solver selection were identical for `SATzilla07`$^+$ and `SATzilla07`$^*$: both chose to first run the local search solver `SAPS` for two CPU seconds, followed by two CPU seconds of `March_ks`. These solvers were similar to our manual selection, but their order was reversed. For the solver subset selection, `SATzilla07`$^+$ and `SATzilla07`$^*$ yielded somewhat different results, but both of them kept two local search algorithms, `Ag2wsat+` & `Ranov`, and `Ag2wsat+` & `Gnovelty+`, respectively.

Table 24 compares the performance of the four versions of `SATzilla` on the `ALL` category. Roughly equal improvements in terms of all our performance metrics were due to more training data and solvers on the one hand, and to the improvements in `SATzilla07`$^+$ on the other hand. The best performance in terms of all our performance metrics was obtained by `SATzilla07`$^*$`(S`$^{++}$`,D`$^+$`)`. Recall that the only difference between `SATzilla07`$^+$`(S`$^{++}$`,D`$^+$`)` and `SATzilla07`$^*$`(S`$^{++}$`,D`$^+$`)` was the use of more general hierarchical hardness models, as described in Section 5.5.

Note that using a classifier is of course not as good as using an oracle for determining the distribution an instance comes from; thus, the success ratios of the solvers selected by `SATzilla07`$^*$ over the instances in the test set for distribution `ALL` (see Table 25) were slightly lower than those for the solvers picked by `SATzilla07`$^+$ for each of the distributions individually (see Tables 16, 19, and 22). However, when compared to `SATzilla07`$^+$ on distribution `ALL`, `SATzilla07`$^*$ performed significantly better: achieving overall performance improvements of 11.3% lower average runtime, 1.8% more solved instances and 9.6% higher score. This supports our initial hypothesis that SATzilla07* would perform slightly worse than specialized versions of SATzilla07$^+$ in each single category, yet would yield the best result when applied to a broader and more heterogeneous set of instances.

| SATzilla | Pre-Solver (time) | Components |
|---|---|---|
| SATzilla07(S,D') | March_dl04(5); SAPS(2) | Eureka, Kcnfs06, March_dl04, Minisat 2.0,Zchaff_rand |
| SATzilla07(S$^+$,D$^+$) | March_dl04(5); SAPS(2) | Eureka, March_dl04, Zchaff_rand, Kcnfs04, TTS, Picosat, March_ks, Minisat07 |
| SATzilla07$^+$(S$^{++}$,D$^+$) | SAPS(2); March_ks(2) | Eureka, Kcnfs06, Rsat 1.03, Zchaff_rand, TTS, MXC, TinisatElite, Rsat 2.0, Ag2wsat+, Ranov |
| SATzilla07*(S$^{++}$,D$^+$) | SAPS(2); March_ks(2) | Eureka, Kcnfs06, March_dl04, Minisat 2.0, Rsat 1.03, Picosat, MXC, March_ks, Minisat07, Ag2wsat+, Gnovelty+ |

Table 23: *SATzilla's configuration for the* ALL *category.*

| Solver | Avg. runtime [s] | Solved [%] | Performance score |
|---|---|---|---|
| Rsat 1.03 | 542 | 61.1 | 131399 |
| Kcnfs04 | 969 | 21.3 | 46695 |
| TTS | 939 | 22.6 | 74616 |
| Picosat | 571 | 57.7 | 135049 |
| March_ks | 509 | 62.9 | 202133 |
| TinisatElite | 690 | 47.3 | 93169 |
| Minisat07 | 528 | 61.8 | 162987 |
| Gnovelty+ | 684 | 43.9 | 156365 |
| March_dl04 | **509** | **62.7** | **205592** |
| SATzilla07(S,D') | 282 | 83.1 | —   (125.0%) |
| SATzilla07(S$^+$,D$^+$) | 224 | 87.0 | —   (139.2%) |
| SATzilla07$^+$(S$^{++}$,D$^+$) | 194 | 91.1 | —   (158%) |
| SATzilla07*(S$^{++}$,D$^+$) | **172** | **92.9** | 344594 (167.6%) |

Table 24: *The performance of* SATzilla *compared to the best solvers on* ALL. *Scores for non-portfolio solvers were computed using a reference set in which the only* SATzilla *solver was* SATzilla07*(S$^{++}$,D$^+$). *Cutoff time:* 1 200 *CPU seconds.*

The runtime cumulative distribution function (Figure 11, right) shows that SATzilla07*(S$^{++}$,D$^+$) dominated the other versions of SATzilla on ALL and solved about 30% more instances than the best non-portfolio solver, March_dl04 (Figure 11, left).

Table 26 shows the performance of the general classifier in SATzilla07*(S$^{++}$,D$^+$). We note several patterns: Firstly, classification performance for RANDOM and INDUSTRIAL in-stances was much better than for `HANDMADE` instances. Secondly, for `HANDMADE` instances, most misclassifications were not due to a misclassification of the type of instance, but rather of the satisfiability status. Finally, we can see that `RANDOM` instances were almost perfectly classified as `RANDOM` and only very few other instances were classified as `RANDOM`, while `HANDMADE` and `INDUSTRIAL` instances were confused somewhat more often. The comparably poor classification performance for `HANDMADE` instances partly explains why `SATzilla07`*`(S`$^{++}$`,D`$^+$`)` did not perform as well for the `HANDMADE` category as for the others.

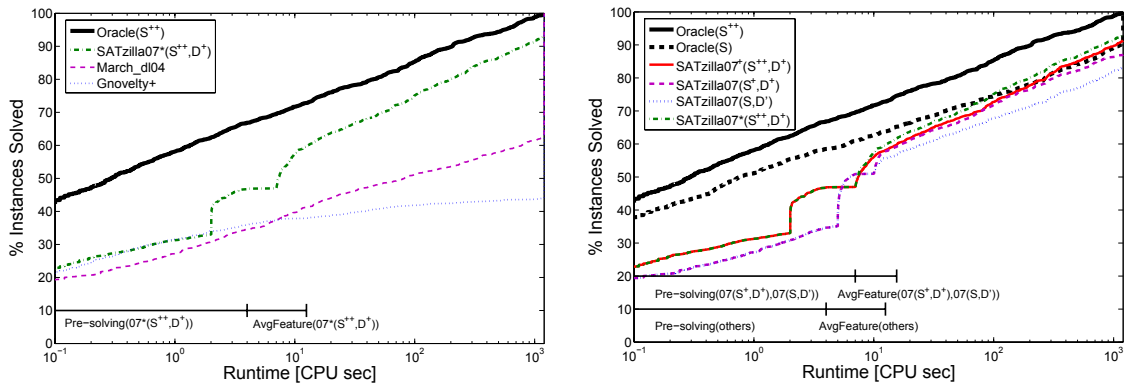

Figure 11: *Left: CDF for* `SATzilla07`*`(S`$^{++}$`,D`$^+$`)` *and the best non-portfolio solvers on* `ALL`*; right: CDFs for different versions of* `SATzilla` *on* `ALL` *shown in Table 23. All other solvers' CDFs are below the ones shown here.*

| Pre-Solver (Pre-Time) | Solved [%] | Avg. Runtime [CPU sec] |
|---|---|---|
| SAPS(2) | 33.0 | 1.4 |
| March_ks (2) | 13.9 | 1.6 |

| Selected Solver | Selected [%] | Success [%] | Avg. Runtime [CPU sec] |
|---|---|---|---|
| Minisat07 | 21.2 | 85.5 | 247.5 |
| March_dl04 | 14.5 | 84.0 | 389.5 |
| Gnovelty+ | 12.5 | 85.2 | 273.2 |
| March_ks | 9.1 | 89.8 | 305.6 |
| Eureka (BACKUP) | 8.9 | 89.7 | 346.1 |
| Eureka | 7.2 | 97.9 | 234.6 |
| Picosat | 6.6 | 90.7 | 188.6 |
| Kcnfs06 | 6.5 | 95.2 | 236.3 |
| MXC | 5.5 | 88.9 | 334.0 |
| Rsat 1.03 | 4.0 | 80.8 | 364.9 |
| Minisat 2.0 | 3.5 | 56.5 | 775.7 |
| Ag2wsat+ | 0.5 | 33.3 | 815.7 |

Table 25: *The solvers selected by* `SATzilla07`*`(S`$^{++}$`,D`$^+$`)` *for the* `ALL` *category.*

|  | R, sat | R, unsat | H, sat | H, unsat | I, sat | I, unsat |
|---|---|---|---|---|---|---|
| classified R, sat | **92%** | 5% | 1% | – | 1% | 1% |
| classified R, unsat | 4% | **94%** | – | 1% | – | 1% |
| classified H, sat | – | – | **57%** | 38% | – | 5% |
| classified H, unsat | – | 1% | 23% | **71%** | 1% | 4% |
| classified I, sat | – | – | 8% | – | **81%** | 11% |
| classified I, unsat | – | – | – | 5% | 6% | **89%** |

Table 26: *Confusion matrix for the 6-way classifier on data set* `ALL`.

## 8. Conclusions

Algorithms can be combined into portfolios to build a whole greater than the sum of its parts. We have significantly extended earlier work on algorithm portfolios for SAT that select solvers on a per-instance basis using empirical hardness models for runtime prediction. We have demonstrated the effectiveness of our general portfolio construction method, `SATzilla07`, on four large sets of SAT competition instances. Our own experiments show that our `SATzilla07` portfolio solvers always outperform their components. Furthermore, `SATzilla07`'s excellent performance in the recent 2007 SAT Competition demonstrates the practical effectiveness of our approach.

In this work, we pushed the `SATzilla` approach further beyond `SATzilla07`. For the first time, we showed that portfolios can optimize complex scoring functions and integrate local search algorithms as component solvers. Furthermore, we showed how to automate the process of pre-solver selection, one of the last aspects of our approach that was previously based on manual engineering. As demonstrated in extensive computational experiments, these enhancements improved `SATzilla07`'s performance substantially.

`SATzilla` is now at a stage where it can be applied "out of the box" given a set of possible component solvers along with representative training and validation instances. In an automated built-in meta-optimization process, the component solvers to be used and the solvers to be used as pre-solvers are automatically determined from the given set of solvers, without any human effort. The computational bottleneck is to execute the possible component solvers on a representative set of instances in order to obtain enough runtime data to build reasonably accurate empirical hardness models. However, these computations can be parallelized very easily and require no human intervention, only computer time, which becomes ever cheaper. Matlab code for building empirical hardness models and C++ code for building `SATzilla` portfolios that use these models are available online at `http://www.cs.ubc.ca/labs/beta/Projects/SATzilla`.

It is interesting to note that the use of local search methods has a significant impact on the performance of `SATzilla`. In preliminary experiments, we observed that the overall performance of `SATzilla07`* was significantly weaker when local search solvers and local-search-based features were excluded. Specifically, the availability of these local search components substantially boosted `SATzilla07*`'s performance on the `RANDOM` instance category and also led to some improvements on `INDUSTRIAL`, but resulted in weaker performance on `HANDMADE` instances. Generally, we believe that a better understanding of the impact of features on our runtime predictions and instance categorizations will allow us to further improve `SATzilla`, and we have therefore begun an in-depth investigation in this direction.

`SATzilla`'s performance ultimately depends on the power of all its component solvers and automatically gets better as they are improved. Furthermore, `SATzilla` takes advantage of solvers that are only competitive for certain kinds of instances and perform poorly otherwise, and thus `SATzilla`'s success demonstrates the value of such solvers. Indeed, the identification of more such solvers, which are otherwise easily overlooked, still has the potential to further improve `SATzilla`'s performance substantially.

## Acknowledgments

This work builds on contributions from a wide range of past co-authors, colleagues, and members of the SAT community. First, we have many colleagues to thank for their contributions to the work described in this article. Eugene Nudelman, Alex Devkar and Yoav Shoham were Kevin and Holger's co-authors on the papers that first introduced `SATzilla` (Nudelman et al., 2004a, 2004b); this work grew out of a project on automated algorithm selection that involved Galen Andrew and Jim McFadden, in addition to Kevin, Eugene and Yoav (Leyton-Brown et al., 2003b, 2003a). Nando de Freitas, Bart Selman, and Kevin Murphy gave useful suggestions about machine learning algorithms, SAT instance features, and mixtures of experts, respectively. Second, while academic research always builds on previous work, we are especially indebted to the authors of the dozens of SAT solvers we discuss in this paper, and particularly to their commitment to furthering scientific understanding by making their code publicly available. Without these researchers' considerable efforts, `SATzilla` could never have been built.

## References

Bacchus, F. (2002a). Enhancing Davis Putnam with extended binary clause reasoning. In *Proceedings of the Eighteenth National Conference on Artificial Intelligence (AAAI'02)*, pp. 613–619.

Bacchus, F. (2002b). Exploring the computational tradeoff of more reasoning and less searching. In *Proceedings of the Fifth International Conference on Theory and Applications of Satisfiability Testing (SAT'02)*, pp. 7–16.

Bacchus, F., & Winter, J. (2003). Effective preprocessing with hyper-resolution and equality reduction. In *Proceedings of the Sixth International Conference on Theory and Applications of Satisfiability Testing (SAT'03)*, pp. 341–355.

Biere, A. (2007). Picosat version 535. Solver description, SAT competition 2007.

Biere, A., Cimatti, A., Clarke, E. M., Fujita, M., & Zhu, Y. (1999). Symbolic model checking using SAT procedures instead of BDDs. In *Proceedings of Design Automation Conference (DAC'99)*, pp. 317–320.

Bishop, C. M. (2006). *Pattern Recognition and Machine Learning*. Springer.

Bregman, D. R., & Mitchell, D. G. (2007). The SAT solver MXC, version 0.5. Solver description, SAT competition 2007.

C. M. Li, W. W., & Zhang, H. (2007). Combining adaptive noise and promising decreasing variables in local search for SAT. Solver description, SAT competition 2007.

Carchrae, T., & Beck, J. C. (2005). Applying machine learning to low-knowledge control of optimization algorithms. *Computational Intelligence*, *21*(4), 372–387.

Crawford, J. M., & Baker, A. B. (1994). Experimental results on the application of satisfiability algorithms to scheduling problems. In *Proceedings of the Twelfth National Conference on Artificial Intelligence (AAAI'94)*, pp. 1092–1097.

Davis, M., Logemann, G., & Loveland, D. (1962). A machine program for theorem proving. *Communications of the ACM*, *5*(7), 394–397.

Davis, M., & Putnam, H. (1960). A computing procedure for quantification theory. *Journal of the ACM*, *7*(1), 201–215.

Dechter, R., & Rish, I. (1994). Directional resolution: The Davis-Putnam procedure, revisited. In *Principles of Knowledge Representation and Reasoning (KR'94)*, pp. 134–145.

Dequen, G., & Dubois, O. (2007). kcnfs. Solver description, SAT competition 2007.

Dubois, O., & Dequen, G. (2001). A backbone-search heuristic for efficient solving of hard 3-SAT formulae. In *Proceedings of the Seventeenth International Joint Conference on Artificial Intelligence (IJCAI'01)*, pp. 248–253.

Eén, N., & Sörensson, N. (2003). An extensible SAT-solver. In *Proceedings of the Sixth International Conference on Theory and Applications of Satisfiability Testing (SAT'03)*, pp. 502–518.

Eén, N., & Sörensson, N. (2006). Minisat v2.0 (beta). Solver description, SAT Race 2006.

Gagliolo, M., & Schmidhuber, J. (2006a). Impact of censored sampling on the performance of restart strategies. In *Twelfth Internatioal Conference on Principles and Practice of Constraint Programming (CP'06)*, pp. 167–181.

Gagliolo, M., & Schmidhuber, J. (2006b). Learning dynamic algorithm portfolios. *Annals of Mathematics and Artificial Intelligence*, *47*(3-4), 295–328.

Gebruers, C., Hnich, B., Bridge, D., & Freuder, E. (2005). Using CBR to select solution strategies in constraint programming. In *Proceedings of the Sixth International Conference on Case-Based Reasoning (ICCBR'05)*, pp. 222–236.

Gebruers, C., Guerri, A., Hnich, B., & Milano, M. (2004). Making choices using structure at the instance level within a case based reasoning framework. In *International Conference on Integration of AI and OR Techniques in Constraint Programming for Combinatorial Optimization Problems (CPAIOR-04)*, pp. 380–386.

Gomes, C. P., & Selman, B. (2001). Algorithm portfolios. *Artificial Intelligence*, *126(1-2)*, 43–62.

Guerri, A., & Milano, M. (2004). Learning techniques for automatic algorithm portfolio selection. In *Proceedings of the 16th European Conference on Artificial Intelligence (ECAI-04)*, pp. 475–479.

Guo, H., & Hsu, W. H. (2004). A learning-based algorithm selection meta-reasoner for the real-time MPE problem. In *Proceedings of the Seventeenth Australian Conference on Artificial Intelligence*, pp. 307–318.

Guyon, I., Gunn, S., Nikravesh, M., & Zadeh, L. (2006). *Feature Extraction, Foundations and Applications*. Springer.

Heule, M., & v. Maaren, H. (2007). march_ks. Solver description, SAT competition 2007.

Heule, M., Zwieten, J., Dufour, M., & Maaren, H. (2004). March_eq: implementing additional reasoning into an efficient lookahead SAT solver. In *Proceedings of the Seventh International Conference on Theory and Applications of Satisfiability Testing (SAT'04)*, pp. 345–359.

Hoos, H. H. (2002). An adaptive noise mechanism for WalkSAT. In *Proceedings of the Eighteenth National Conference on Artificial Intelligence (AAAI'02)*, pp. 655–660.

Hoos, H. H., & Stützle, T. (2005). *Stochastic Local Search - Foundations & Applications*. Morgan Kaufmann Publishers, San Francisco, CA, USA.

Horvitz, E., Ruan, Y., Gomes, C. P., Kautz, H., Selman, B., & Chickering, D. M. (2001). A Bayesian approach to tackling hard computational problems. In *Proceedings of the Seventeenth Conference on Uncertainty in Artificial Intelligence (UAI'01)*, pp. 235–244.

Huang, J. (2007). TINISAT in SAT competition 2007. Solver description, SAT competition 2007.

Huberman, B., Lukose, R., & Hogg, T. (1997). An economics approach to hard computational problems. *Science*, *265*, 51–54.

Hutter, F., Hamadi, Y., Hoos, H. H., & Leyton-Brown, K. (2006). Performance prediction and automated tuning of randomized and parametric algorithms. In *Twelfth Internatioal Conference on Principles and Practice of Constraint Programming (CP'06)*, pp. 213–228.

Hutter, F., Tompkins, D. A. D., & Hoos, H. H. (2002). Scaling and probabilistic smoothing: Efficient dynamic local search for SAT. In *Proceedings of the Eighth International Conference on Principles and Practice of Constraint Programming*, pp. 233–248.

Ishtaiwi, A., Thornton, J., Anbulagan, Sattar, A., & Pham, D. N. (2006). Adaptive clause weight redistribution. In *Twelfth Internatioal Conference on Principles and Practice of Constraint Programming (CP'06)*, pp. 229–243.

Kautz, H., & Selman, B. (1996). Pushing the envelope: Planning, propositional logic, and stochastic search. In *Proceedings of the Thirteenth National Conference on Artificial Intelligence and the Eighth Innovative Applications of Artificial Intelligence Conference*, pp. 1194–1201.

Kautz, H. A., & Selman, B. (1999). Unifying SAT-based and graph-based planning. In *Proceedings of the Sixteenth International Joint Conference on Artificial Intelligence (IJCAI'99)*, pp. 318–325.

Knuth, D. (1975). Estimating the efficiency of backtrack programs. *Mathematics of Computation, 29*(129), 121–136.

Krishnapuram, B., Carin, L., Figueiredo, M., & Hartemink, A. (2005). Sparse multinomial logistic regression: Fast algorithms and generalization bounds. In *IEEE Transactions on Pattern Analysis and Machine Intelligence*, pp. 957–968.

Kullmann, O. (2002). Investigating the behaviour of a SAT solver on random formulas. http://cs-svr1.swan.ac.uk/∼csoliver/Artikel/OKsolverAnalyse.html.

Lagoudakis, M. G., & Littman, M. L. (2001). Learning to select branching rules in the DPLL procedure for satisfiability. In *LICS/SAT*, pp. 344–359.

Le Berre, D., & Simon, L. (2004). Fifty-five solvers in Vancouver: The SAT 2004 competition. In *Proceedings of the Seventh International Conference on Theory and Applications of Satisfiability Testing (SAT'04)*, pp. 321–344.

Leyton-Brown, K., Nudelman, E., Andrew, G., McFadden, J., & Shoham, Y. (2003a). Boosting as a metaphor for algorithm design. In *Ninth Internatioal Conference on Principles and Practice of Constraint Programming (CP'03)*, pp. 899–903.

Leyton-Brown, K., Nudelman, E., Andrew, G., McFadden, J., & Shoham, Y. (2003b). A portfolio approach to algorithm selection. In *Proceedings of the Eighteenth International Joint Conference on Artificial Intelligence (IJCAI'03)*, pp. 1542–1543.

Leyton-Brown, K., Nudelman, E., & Shoham, Y. (2002). Learning the empirical hardness of optimization problems: The case of combinatorial auctions. In *Eighth Internatioal Conference on Principles and Practice of Constraint Programming (CP'02)*, pp. 556–572.

Li, C., & Huang, W. (2005). Diversification and determinism in local search for satisfiability. In *Proceedings of the Eighth International Conference on Theory and Applications of Satisfiability Testing (SAT'05)*, pp. 158–172.

Lobjois, L., & Lemaître, M. (1998). Branch and bound algorithm selection by performance prediction. In *Proceedings of the Fifteenth National Conference on Artificial Intelligence (AAAI'98)*, pp. 353–358.

Mahajan, Y. S., Fu, Z., & Malik, S. (2005). Zchaff2004: an efficient SAT solver. In *Proceedings of the Eighth International Conference on Theory and Applications of Satisfiability Testing (SAT'05)*, pp. 360–375.

Murphy, K. (2001). The Bayes Net Toolbox for Matlab. In *Computing Science and Statistics*, Vol. 33. http://bnt.sourceforge.net/.

Nadel, A., Gordon, M., Palti, A., & Hanna, Z. (2006). Eureka-2006 SAT solver. Solver description, SAT Race 2006.

Nudelman, E., Leyton-Brown, K., Hoos, H. H., Devkar, A., & Shoham, Y. (2004a). Understanding random SAT: Beyond the clauses-to-variables ratio. In *Tenth Internatioal Conference on Principles and Practice of Constraint Programming (CP'04)*, pp. 438–452.

Nudelman, E., Leyton-Brown, K., Devkar, A., Shoham, Y., & Hoos, H. (2004b). Satzilla: An algorithm portfolio for SAT. Solver description, SAT competition 2004.

Pham, D. N., & Anbulagan (2007). Resolution enhanced SLS solver: R+AdaptNovelty+. Solver description, SAT competition 2007.

Pham, D. N., & Gretton, C. (2007). gNovelty+. Solver description, SAT competition 2007.

Pipatsrisawat, K., & Darwiche, A. (2006). Rsat 1.03: SAT solver description. Tech. rep. D-152, Automated Reasoning Group, UCLA.

Pipatsrisawat, K., & Darwiche, A. (2007). Rsat 2.0: SAT solver description. Solver description, SAT competition 2007.

Rice, J. R. (1976). The algorithm selection problem. *Advances in Computers*, *15*, 65–118.

Samulowitz, H., & Memisevic, R. (2007). Learning to solve QBF. In *Proceedings of the Twentysecond National Conference on Artificial Intelligence (AAAI'07)*, pp. 255–260.

Schmee, J., & Hahn, G. J. (1979). A simple method for regression analysis with censored data. *Technometrics*, *21*(4), 417–432.

Selman, B., Kautz, H., & Cohen, B. (1994). Noise strategies for improving local search. In *Proceedings of the Twelfth National Conference on Artificial Intelligence (AAAI'94)*, pp. 337–343.

Selman, B., Levesque, H., & Mitchell, D. (1992). A new method for solving hard satisfiability problems. In *Proceedings of the Tenth National Conference on Artificial Intelligence (AAAI'92)*, pp. 440–446.

Sörensson, N., & Eén, N. (2007). Minisat2007. http://www.cs.chalmers.se/Cs/Research/FormalMethods/MiniSat/.

Spence, I. (2007). Ternary tree solver (tts-4-0). Solver description, SAT competition 2007.

Stephan, P., Brayton, R., & Sangiovanni-Vencentelli, A. (1996). Combinational test generation using satisfiability. *IEEE Transactions on Computer-Aided Design of Integrated Circuits and Systems*, *15*, 1167–1176.

Streeter, M., Golovin, D., & Smith, S. F. (2007). Combining multiple heuristics online. In *Proceedings of the Twentysecond National Conference on Artificial Intelligence (AAAI'07)*, pp. 1197–1203.

Subbarayan, S., & Pradhan, D. (2005). Niver: Non-increasing variable elimination resolution for preprocessing sat instances. *Lecture Notes in Computer Science, Springer*, *3542/2005*, 276–291.

Tompkins, D. A. D., & Hoos, H. H. (2004). UBCSAT: An implementation and experimentation environment for SLS algorithms for SAT & MAX-SAT.. In *Proceedings of the Seventh International Conference on Theory and Applications of Satisfiability Testing (SAT'04)*.

Vallstrom, D. (2005). Vallst documentation. http://vallst.satcompetition.org/index.html.

van Gelder, A. (2002). Another look at graph coloring via propositional satisfiability. In *Proceedings of Computational Symposium on Graph Coloring and Generalizations (COLOR-02)*, pp. 48–54.

Wei, W., Li, C. M., & Zhang, H. (2007). Deterministic and random selection of variables in local search for SAT. Solver description, SAT competition 2007.

Xu, L., Hoos, H. H., & Leyton-Brown, K. (2007a). Hierarchical hardness models for SAT. In *Thirteenth Internatioal Conference on Principles and Practice of Constraint Programming (CP'07)*, pp. 696–711.

Xu, L., Hutter, F., Hoos, H., & Leyton-Brown, K. (2007b). Satzilla-07: The design and analysis of an algorithm portfolio for SAT. In *Thirteenth Internatioal Conference on Principles and Practice of Constraint Programming (CP'07)*, pp. 712–727.

Xu, L., Hutter, F., Hoos, H., & Leyton-Brown, K. (2007c). Satzilla2007: a new & improved algorithm portfolio for SAT. Solver description, SAT competition 2007.

Zhang, L., Madigan, C. F., Moskewicz, M. W., & Malik, S. (2001). Efficient conflict driven learning in Boolean satisfiability solver. In *Proceedings of the International Conference on Computer Aided Design*, pp. 279–285.

Zhang, L. (2002). The quest for efficient Boolean satisfiability solvers. In *Proceedings of 8th International Conference on Computer Aided Deduction (CADE-02)*, pp. 313–331.